\documentclass{article}

\usepackage{arxiv}
\usepackage{adjustbox}
\usepackage{blindtext}
\usepackage[utf8]{inputenc} 
\usepackage[T1]{fontenc}    
\usepackage{hyperref}       
\usepackage{url}            
\usepackage{booktabs}       
\usepackage{amsfonts}       
\usepackage{nicefrac}       
\usepackage{lipsum}		
\usepackage{graphicx}
\usepackage{doi}
\usepackage[table, svgnames]{xcolor}
\usepackage{multirow}
\usepackage{geometry}
\usepackage{float}
\usepackage[table, svgnames]{xcolor}
\usepackage{tabularx, caption, makecell, booktabs,enumitem}
\usepackage[utf8]{inputenc}
\usepackage{tikz,float}
\usepackage{tabularx}
\usepackage{xparse}

\usepackage{amsmath,amssymb,amsthm}
\usepackage{rotating}

\title{Vision Transformers for Action Recognition: A Survey}

\author{
  Anwaar Ulhaq \\
  Charles Sturt University, \\
  Port Macquarie, NSW, Australia, 2444 \\
   \texttt{aulhaq@csu.edu.au} \\
   \And
 Naveed Akhtar \\
University of Western Austrlia \\
 Perth, WA, Australia, 
 6000\\
  \texttt{naveed.akhtar@uwa.edu.au} \\
 \AND
  Ganna Pogrebna\\
 Charles Sturt University, \\
  Port Macquarie, NSW, Australia, 2444 \\
   \texttt{\{gpogrebna@csu.edu.au} \\
  \And
  Ajmal Mian\\
University of Western Austrlia \\
 Perth, WA, Australia,
  6000\\
  \texttt{ajmal.mian@uwa.edu.au} \\
}
\NewDocumentCommand{\statcirc}{ O{#2} m }{%
    \begin{tikzpicture}
    \fill[#2] (0,0) circle (1.0ex); 
    \fill[#1] (0,0) -- (180:1ex) arc (180:0:1ex) -- cycle; 
    \end{tikzpicture}
}

\renewcommand{\shorttitle}{\textit{arXiv} }
\hypersetup{
pdftitle={A template for the arxiv style},
pdfsubject={q-bio.NC, q-bio.QM},
pdfauthor={David S.~Hippocampus, Elias D.~Striatum},
pdfkeywords={First keyword, Second keyword, More},
}
\begin{document}
\maketitle

\begin{abstract}
Vision transformers are emerging as a powerful tool to solve computer vision problems. Recent techniques have also proven the efficacy of transformers beyond the image domain to solve numerous video-related tasks. Among those, human action recognition is receiving special attention from the research community due to its widespread applications. This article provides the first comprehensive survey of vision transformer techniques for action recognition. We analyze and summarize the existing and emerging literature in this direction while highlighting the popular trends in adapting transformers for action recognition. Due to their specialized application, we collectively refer to these methods as ``action transformers''. Our literature review provides suitable taxonomies for action transformers based on their architecture, modality, and intended objective. Within the context of action transformers, we explore the techniques to encode spatio-temporal data,  dimensionality reduction, frame patch and spatio-temporal cube construction, and various representation methods. We also investigate the optimization of  spatio-temporal attention in transformer layers to handle longer sequences, typically by reducing the number of tokens in a single attention operation. Moreover, we also investigate different network learning strategies, such as self-supervised and zero-shot learning, along with their associated losses for transformer-based action recognition. This survey also summarizes the progress towards gaining grounds on evaluation metric scores on important benchmarks with action transformers. Finally, it provides a discussion on the challenges, outlook, and future avenues for this research direction.

\end{abstract}


\section{Introduction}

The huge success of transformers in processing sequential data, especially in natural language processing (NLP), has led to the development of vision transformers that outperform Convolutional neural networks for image recognition tasks on large datasets such as Imagenet \cite{SurveyT1, SurveyT2, SurveyT3}. It has created a paradigm shift in the architecture of neural network models for image recognition tasks. Video recognition - unlike image recognition - solves the problem of event recognition in video sequences, such as human action recognition. Video transformer models have emerged as attractive and promising solutions for improving the accuracy of challenging video recognition tasks such as action recognition. This survey paper provides a comprehensive review of the most recent work related to vision transformer-based action recognition.    

Recognition of human actions is the task of classifying which action is being performed in a video. Action detection and segmentation, on the other hand, address localization or extraction of action instances from videos. To learn accurate representations of actions, deep learning models typically need access to enormous video datasets which makes it challenging given the high dimensionality and volume of video data. In addition, the ability of these models to extract spatial and temporal complexity inside video representations is crucial for action recognition. Over the last ten years, action recognition has been extensively studied, and various review articles and survey papers are available in general. However,  most of these review papers \cite{RecentReview, ModalitySurvey, survey2022} focus on convolutional neural networks or traditional machine learning models for action recognition. As transformer architecture is driving a new paradigm shift in computer vision and researchers are rapidly adapting transformer architectures to improve the accuracy and efficiency of action recognition tasks. Hence, a recent review of the state-of-the-art research is necessary to steer this research direction. This review paper has an entirely different focus on putting transformer architecture as an emerging model for action recognition.

\begin{figure}[t]
    \centering
    \includegraphics[width = 0.97\textwidth]{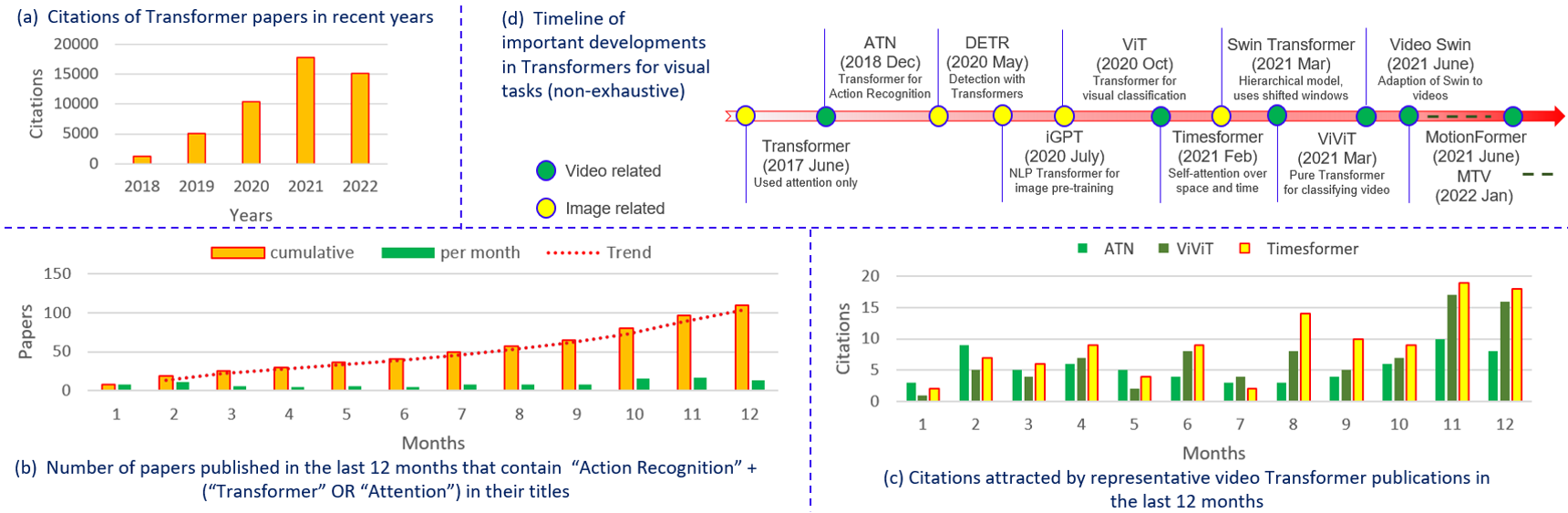}
    \caption{Statistics and timeline overview of the literature on Transformers. (a) Number of per-year Google Scholar citations attracted by Transformer papers. (b) Trend of the number of papers published in Action Recognition using attention mechanism in the last 12 months. (c) Citations attracted by  ATN~\cite{VATN}, ViViT~\cite{vivit}, and Timesformer~\cite{Timeformer} in the last 12 months as representative video Transformer papers. (d) A non-exhaustive timeline summary of prominent developments in Transformers for visual tasks.}
    \label{fig:graphs}
\end{figure}

The sole concept that is behind this transformation is attention which is arguably one of the most powerful concepts in deep learning today. The idea that when processing a large amount of data, we ``attend to" a specific component is grounded in the process of human cognition. Considering that textual data is typically presented sequentially, the NLP was the first field to propose the attention mechanism. The context vector was proposed by Bahdanau et al. \cite{Bahdanaul} to connect the source's and target sequence inputs. When an encoder cell's hidden state is aligned with the current target output, the encoder cell's context vector is preserved. By doing so, the model can ``attend to" a portion of the source inputs and learn the complex source-target relationship more effectively. Apart from machine translation, the concept of attention was first used for image captioning. Xu et al.\cite{XU} proposed an attention framework that surpasses the Seq2Seq architecture. Their framework tries to align the input image with the output word to solve the problem of image captioning. Subsequently, they used a convolutional layer to extract image features and an RNN with Attention to align these features captions that are automatically generated to draw attention to specific details in the image. The generated words (captions) are positioned so that they draw attention to the objects of interest in the image. With their framework, they were one of the first to try to solve problems outside of neural machine translation.

The most significant attention-based model transformer was proposed by Vaswani et al. \cite{Transformer}. With Transformer's multi-head self-attention layer, a representation of a sequence can be computed by aligning words within the sequence with other words within the sequence. Compared to convolution and recursive operations, it performs better in representation and uses fewer computational resources. To that end, the Transformer architecture forgoes convolution and recursive procedures in favor of more focused use of multiple processing nodes. When using multi-head attention, multiple attention layers work together to learn a variety of representations from various viewpoints. Several natural language processing tasks have achieved human-level performance by using models inspired by Transformers, such as GPT \cite{GPT3}and BERT \cite{Bert}.

The success of the transformer models inspired the computer vision community to test it for pure computer vision tasks. The work by Dosovitskiy et al. \cite{VIT} was one of the first models that successfully demonstrated that a Transformer could be used for computer vision tasks. Furthermore, when trained with sufficient data, Vision Transformer outperforms state-of-the-art, large-scale CNN models \cite{CNN}.  This could indicate that the golden age of CNN in computer vision, which lasted years, will be overshadowed by Vision Transformer models, just as it was with  RNN's \cite{RNN} .

Video recognition tasks were logically the next challenge for vision transformers. Video recognition - unlike image recognition - solves the problem of event recognition in video sequences, such as human action recognition. Many human action recognition models had proposed integrating attention mechanisms with convolutional and recurrent blocks to improve model accuracy \cite{survey2022}. However, over time, pure transformer architecture-based action recognition has started emerging with an increasing trend. These approaches vary in their design in terms of encoding, attention, and attention mechanism, with different performances in terms of efficiency and accuracy on diverse datasets. To better design vision transformer-based action recognition approaches, researchers require a single, comprehensive but refined source of information as a review or survey. This paper is intended to fill up this gap by searching, investigating, and classifying the action recognition literature based on the vision transformer backbone.

\textbf{Motivation}: Human actions usually involve long-term space-time interactions. Attention mechanisms suppress redundant information and model long-range interactions in space and time in a better way. This is the reason that transformers consistently reach state-of-the-art accuracy in video action recognition. Recent transformer-based models show much promise in dealing with the complexity of video data comprising human actions. However, the field is rapidly evolving, and new challenges are being highlighted. For instance, a recent evaluation \cite{Evaluating} of action transformers versus CNN-based action recognition networks shows that such models are poor in motion modeling and currently incur too much latency overhead. As the field is rapidly evolving and action recognition researchers are interested to know the strengths and pitfalls of vision transformers for action recognition, a comprehensive survey is required that provide them with insights from the existing works and motivate them to address limitations and challenges regarding action transformers. In the absence of such a detailed study, we are motivated to write this comprehensive survey about vision transformers for action recognition to drive future research in this direction. We hope that this survey will provide a one-stop shop description of the existing work and future directions in the field.

\textbf{Contributions}: This review paper is, to the best of our knowledge, the first of its kind to comprehensively examine the recent development of vision transformers for action recognition. This paper aims to provide a systematic categorization of the most recent visual Transformers for action recognition tasks and provide their comprehensive review.

(1) Comprehensiveness: This paper comprehensively reviews over one hundred 
action Transformers 
according to their applications on four fundamental action recognition tasks (i.e., classification, detection,  segmentation, and anticipation),  data modalities (i.e., uni-modal and multimodal data), and three types of dominant architectures (CNN based Transformer, Transformer based CNN and Pure Transformer model). We select more representative methods to give their detailed descriptions and analysis, while introducing other related works in a condensed manner. Not only do we conduct an exhaustive analysis of each model from a single view point, but we also construct their internal connections in terms of efficiency, scalability, explainability, and robustness.

(2) Intuitive Comparison. Due to the fact that existing action Transformers use distinct training schemes and hyper-parameter settings for action recognition tasks, this survey provides multiple lateral comparisons across distinct datasets and constraints. Importantly, we summarise a set of exciting constituents designed for each task, including (a) tokenizing approaches, (b) self-attention and cross-attention mechanisms, and (c) frameworks for task classification.

(3) Comprehensive Analysis:  In addition, we offer well-considered insights from the following perspectives: (a) thematic analysis of existing literature; and (b) automated clustering and visualization of existing literature. We conclude by outlining future research directions.

The structure of the rest of the paper is as follows:  In section 2, we provide a historical development and evolution of transformer architecture. In section 3, we discuss various taxonomies related to action transformers. In section 4, we discuss the performance of existing methods, discuss the bibliographical analysis and provide a visualization of action transformer literature. Section 5 provides a discussion and future directions, followed by a conclusion and references.

\section{Evolution of Transformers from NLP to Computer Vision }
In this section, we discuss the evolutionary development of Transformers and their extensions for visual tasks. Our discussion is instructive, in that it provides an easily understandable description of the key concepts of Transformers while providing the historical account. We directly focus on the central ideas while explaining the concepts for conciseness.

\subsection{The Birth of Transformers in NLP}
\begin{figure}[t]
    \centering
    \includegraphics[width =0.9\textwidth]{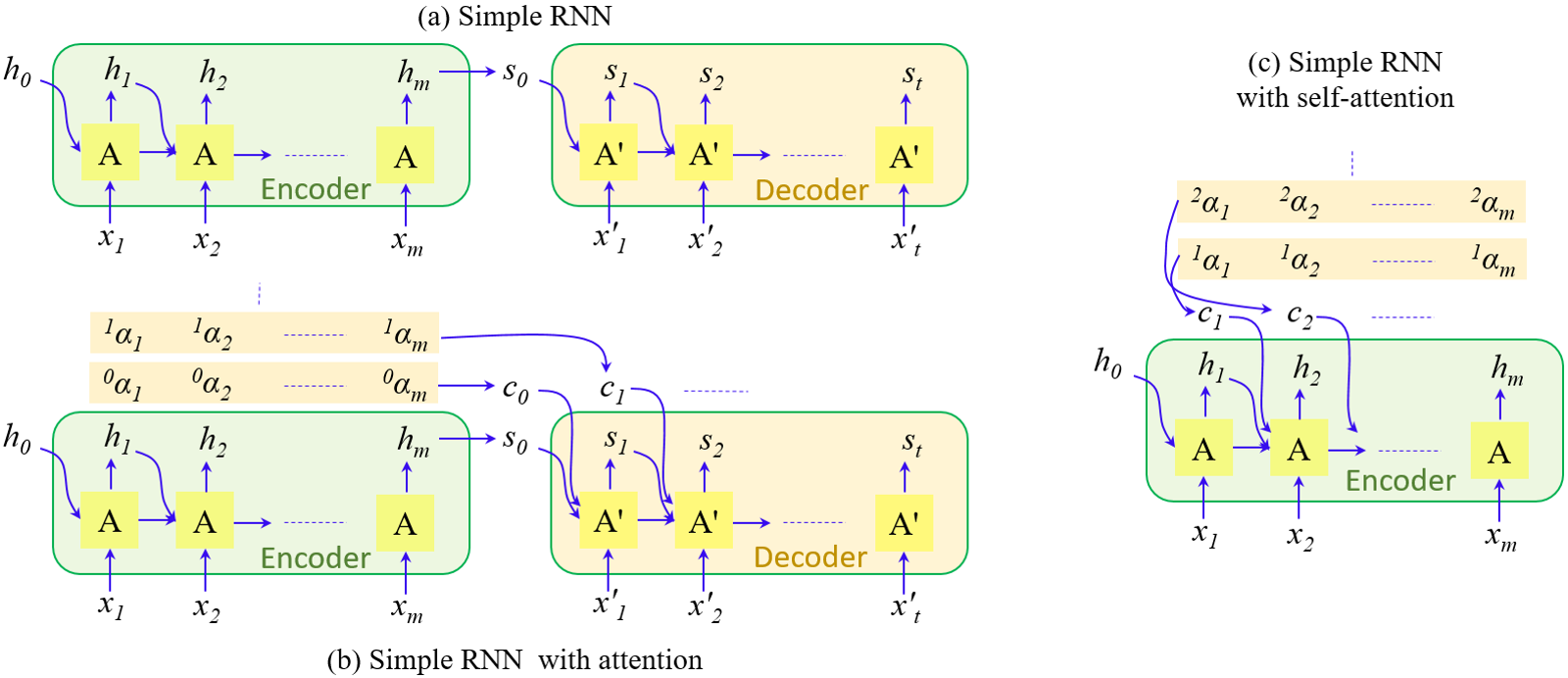}
    \caption{\textbf{(a)}: A simple RNN encoder-decoder model, encoding a sequence of $m$ inputs and decoding $t$ entities. The decoder hidden states $s_j$ are influenced by the immediate previous states. \textbf{(b)}: Attention mechanism computes context vectors $c_j$ that encode the similarity of a state $s_j$ with all the encoder states $h_i, \forall i \in \{1,...,m\}$ through weights $^j\alpha_i$. This allows the decoder states to attend to the encoder states according to their relevance. \textbf{(c)}: Self-attention allows each state of the encoder to attend to all of its own states with the context vectors $c_i$. Self-attention does not involve a decoder. For both attention and self-attention, a weight $^j\alpha_i$ is computed with all \textit{keys} and a \textit{query} - encoding similarities of a \textit{query} with all \textit{keys}. The \textit{values} are combined with these weights to compute the context vector. }
    \label{fig:simpleRNN}
\end{figure}

Recurrent Neural Networks (RNNs) \cite{lipton2015critical} 
are a popular tool to model sequential data. However, their basic scheme has a weakness in that the models tend to forget earlier inputs in a long sequence. For instance, a simple RNN-based encoder-decoder model, e.g.~Figure~\ref{fig:simpleRNN}(a),  can be used for a sequence-to-sequence modeling task such as machine translation~\cite{yang2020survey}. 
Here, encoding is done for a  sequence with `$m$' time stamps, and decoding is performed for `$t$' time stamps. Notice that, a hidden state $h_i$ of the encoder is influenced only by the previous hidden state $h_{i-1}$ and the same is true for the decoder. This can cause the final hidden state $h_m$ to forget the information from the earlier states when $m$ is large. Analogously, the decoder also suffers from the same problem. Moreover, the input state for the decoder, i.e., $s_o$ is $h_m$, which is already sub-optimal for large values of $m$. 
Note that, in this example and the other models shown in Fig.~\ref{fig:simpleRNN}, we ignore details to focus only on the relevant concepts.

The problem of forgetting in RNNs motivated researchers to devise a mechanism that allows the model to pay more \textit{attention} to the important states of the sequence so that they are not forgotten. Whereas \textit{attention}  nowadays is often used as a synonym for  Transformers, in the pioneering work, Bahanaul~\cite{Bahdanaul} devised the attention mechanism in the context of RNNs. The key concept of this mechanism is illustrated in Fig.~\ref{fig:simpleRNN}(b). Primarily, attention computes a \textit{context vector} and uses that, along with the previous state, to compute the next state of the decoder. A context vector $c_j$ governs the attention paid by the decoder to the encoder states. In a sense, this vector accounts for the similarities between the encoder states $h_i, \forall i \in \{1,...,m\}$ and the decoder state $s_j$. There can be multiple ways to compute the similarities. In the wake of Transformers, this is generally done by first computing  \textit{keys} $\boldsymbol{k}_i = \boldsymbol{W}_K h_i$  and \textit{queries} $\boldsymbol{q}_j = \boldsymbol{W}_Q s_j$, where $\boldsymbol W_K$  and $\boldsymbol W_Q$ are learnable transformation matrices. The similarity between a key and a query can then be computed as $^j\alpha_i = \boldsymbol k_i^{\intercal} \boldsymbol q_j$. The context vector can then be computed as $c_j = \sum_{i = 1}^{m}{^j\alpha_i \boldsymbol v_i} $, where $\boldsymbol{v}_i = \boldsymbol W_V h_i$. The vector $\boldsymbol v_i$ is known as \textit{value} and $\boldsymbol W_V$ is also a learnable matrix. Notice that, through $^j\alpha_i$ and $\boldsymbol v_i$ ($\forall i$), all encoder states can influence every decoder state with the help of the context vectors $c_j$. The ability of the decoder states to 
take into consideration all encoder states is the key strength of the attention mechanism.   

Chen et al. \cite{chen2020generative} tried to remove this redundancy by ignoring $\boldsymbol A, \boldsymbol A', h_i$ and $s_j$ altogether, and computing the attention directly with $\boldsymbol W_Q, \boldsymbol W_K$ and $\boldsymbol W_V$.     
Vaswani et al.~\cite{Transformer}  estimate  $\boldsymbol{W}_K, \boldsymbol{W}_Q$ and $\boldsymbol{W}_V$ over the input sequence to generate the context vectors. In doing so, the model looks at the whole sequence simultaneously to compute the context vectors. This removes the recurrence from sequence modeling and gives rise to the architectures known as Transformers. As noted above, we skip over many details to focus only on the central concepts here. Interested readers are recommended a deep dive into the original papers for more details.

\subsection{The Rise of Transformers in Computer Vision}
\begin{figure}[t]
    \centering
    \includegraphics[width = 0.8\textwidth]{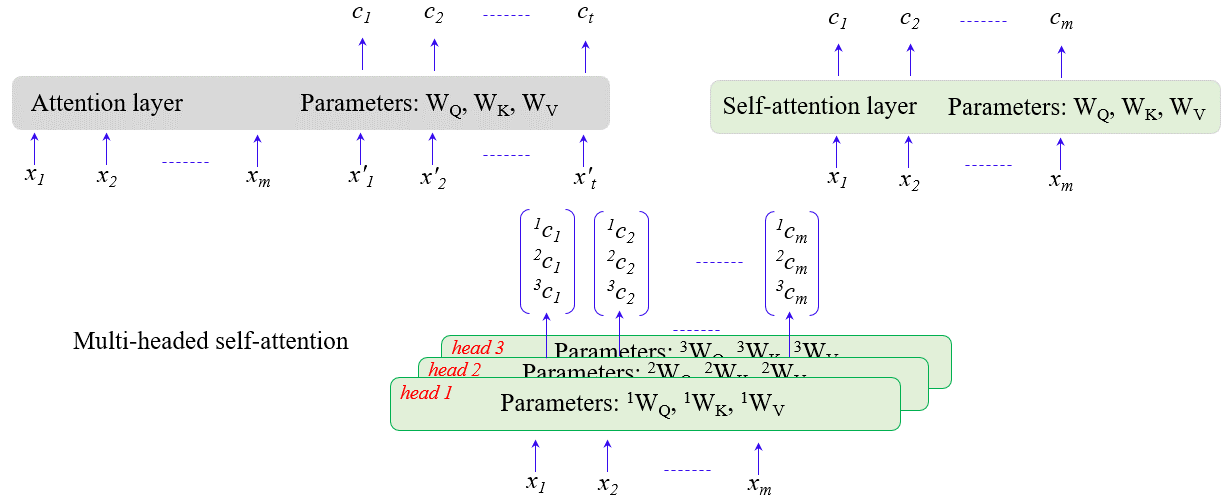}
    \caption{The attention layer computes context vectors for sequence-to-sequence modeling. Self-attention layer can be used to create multiple heads. The context vectors of each head are different due to different initialization of $\boldsymbol Ws$ in each head. The outputs of heads are concatenated for multi-head attention.}
    \label{fig:Layers}
\end{figure}

Whereas examples of employing Transformers in vision tasks can also be found in earlier works, e.g.,\cite{chen2020generative},  
the rise of Transformers in computer vision is mainly due to \cite{VIT}, where Dosovitskiy et al.~successfully adapted the encoder of \cite{Transformer} for the visual classification task. To explain this adaption, we first abstract away the details from the previous section and pack the underlying operations in network layers. In Fig.~\ref{fig:Layers}, the \textit{Attention layer} has learnable parameters $\boldsymbol W_Q, \boldsymbol W_K$ and $\boldsymbol W_V$ that compute the context vectors for a sequence $\{x'_j\}_{j = 1}^{t}$ using the sequence $\{x_i\}_{i = 1}^{m}$. This layer relates to encoder-decoder-based sequence-to-sequence modeling, hence it is less relevant to the vision tasks. The analogous \textit{Self-attention} layer is of key importance in the Transformers in computer vision. Outputs of parallel self-attention layers with their matrices, e.g., $^p\boldsymbol{W}_Q$, $^p\boldsymbol{W}_K$, $^p\boldsymbol{W}_V$, can be concatenated to form \textit{Multi-headed self-attention}, where $p$ denotes the total heads of the attention mechanism. In computer vision literature, it is common to refer to \textit{self-attention} as just \textit{attention} because sequence-to-sequence modeling is seldom required in the field.

\begin{figure}[t]
    \centering
    \includegraphics[width = 0.9\textwidth]{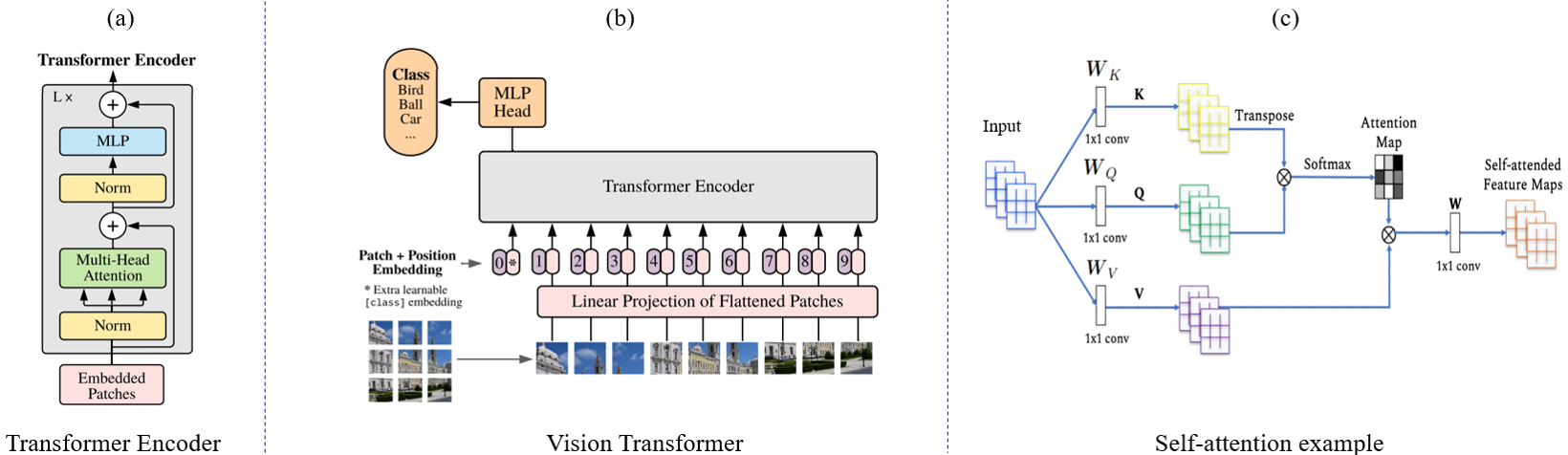}
    \caption{(a) Encoder of Vision Transformer (ViT)~\cite{VIT} inspired by the encoder of \cite{Transformer}. (b) Vision Transformer for classification~\cite{VIT}. (c) An example of self-attention computation in vision tasks~\textcolor{red}{[XX]}. 
    (a) \& (b) are adapted from \cite{VIT} and (c) is adapted from \cite{SurveyT1}.
    } 
    \label{fig:Vit}
\end{figure}

Figure~\ref{fig:Vit}(a) shows the encoder used by the Vision Transformer (ViT)~\cite{VIT}. The encoder is taken from the encoder-decoder Transformer of \cite{Transformer} 
and Multi-Head Attention (MHA) is the  Transformer-related module. The Multi-Layer Perceptron (MLP), Normalization (Norm),  Skip-connections, and addition operation are the standard deep learning components. Conceptually, the  MHA is the same as the multi-headed self-attention discussed in Figure~\ref{fig:Layers}. Figure~\ref{fig:Vit}(b) shows the process of leveraging the Transformer Encoder for visual classification. The input image is first broken into patches which are vectorized (flattened) and projected onto an embedding subspace - similar to computing word embeddings in NLP. A positional embedding is then added to the embedded patches, a.k.a.~tokens, to incorporate the spatial consistency in the learned representation. The ViT uses an additional class token for classification purposes. The context vector of this input is classified by an MLP-based classifier (MLP head).    

In the vision literature, the attention mechanism is frequently described by the following or similar expression 
\begin{equation}
    \text{Attention}(\boldsymbol Q, \boldsymbol K, \boldsymbol V) = \boldsymbol V \cdot ~\text{softmax} \left( \frac{\boldsymbol Q \boldsymbol K^{\intercal}}{\sqrt{d_k}}  \right).
    \label{eq:att}
\end{equation}
The above equation actually performs the context vector computation, as described in the previous section. Here, $\boldsymbol Q$, $\boldsymbol K$ and $\boldsymbol V$ pack the \textit{query}, \textit{key} and \textit{value} vectors into matrices. The  `$\text{softmax} (\frac{\boldsymbol Q \boldsymbol K^{\intercal}}{\sqrt{d_k}})$' term computes the desired $\alpha$'s, where $\sqrt{d_k}$ ($d_k$ is the dimensionality of the \textit{key} vectors) is used for gradient stability and softmax(.) is applied to convert $\alpha$'s into probability distributions. To clarify, in Fig.~\ref{fig:simpleRNN}, the softmax will have the effect of $\sum_{i=1}^{m} {^j}\alpha_i = 1$. 

Whereas Eq.~(\ref{eq:att}) computes the attention - context-vectors to be more precise, recall that computing keys, queries and values requires $\boldsymbol W_K, \boldsymbol W_Q$ and $\boldsymbol W_V$. These parameters can be learned in different ways for vision tasks. For instance, in Fig.~\ref{fig:Vit}(c), we illustrate the self-attention mechanism of \cite{zhang2019self} 
where these parameters are weights of $1\times 1$ convolutional layers. The overall graph implements Eq.~(\ref{eq:att}) with the exceptions of ignoring the scaling by $\sqrt{d_k}$ and post-processing the attention (map) with another $1\times 1$ convolutional  layer, having weights $\boldsymbol W$.  

After impressive classification results achieved by ViT~\cite{VIT}, Transformers have attracted significant interest from the computer vision community~\cite{SurveyT1}. For all the mainstream vision tasks, e.g., recognition, detection, and generation, Transformers are now replacing or augmenting convolutional neural models. The inclination of the vision community towards Transformers is not only due to the impressive performance of this technology, it is also because Transformers have a lower vision-specific inductive bias. This allows for learning more effective models 
from 
more training data.    

\subsection{Transformers for Videos}
Videos are basically image sequences and, hence, have a natural affinity to Transformers. However, the application of Transformers to videos resembles more their application to images~\cite{VIT} than to the original notion of sequence-to-sequence modeling~\cite{Transformer}. Early examples of  Transformers in the video domain include Timesformer \cite{Timeformer} and  ViViT \cite{vivit} (for Video Vision Transformer). Whereas more details of these and other architectures will come in the later part of the article, we briefly summarize a few relevant aspects here.

To model video clips, video Transformers propose novel embedding schemes and variations to ViT~\cite{VIT} and related Transformers. Similar to the image input of ViT, frames in videos can be tokenized by being split up into smaller patches and then flattened in space. For instance, the authors of Timesformer \cite{Timeformer} propose a tokenization scheme called uniform frame sampling, wherein randomly selected frames from a clip are used for tokenization. 

In contrast to \cite{Timeformer}, ViViT \cite{vivit} introduced Tubelets Embedding that can preserve contextual time data in the video. As a first step, the model extracts volumes from the video, which includes both frame and time stamp patches. Afterward, the volumes are flattened to generate tokens. The authors propose four distinct variants of their method, called Spatio-temporal attention, Factorized Encoder, Factorized self-attention, and Factorized dot-product attention based on the underlying techniques. 

 We use ViViT \cite{vivit}  as a representative for video transformers instead of Timesformer \cite{Timeformer}  as it works on 3D volumes instead of frames. It extracts non-overlapping, spatio-temporal ``tubes" from the input volume and applies linear projection. This technique is analogous to 3D convolution and represents a 3D extension of ViT's embedding. Tokens are taken from the height, width, and time dimensions of a tubelet. Therefore, more tokens are needed when the tubelet dimensions are reduced. This technique differs from the above-noted uniform frame sampling, in which the transformer fuses temporal information across frames by fusing spatial and temporal details during tokenization.

\section{Visual Action Transformers: Taxonomies}
\begin{figure}
\centering
\includegraphics[width=100mm]{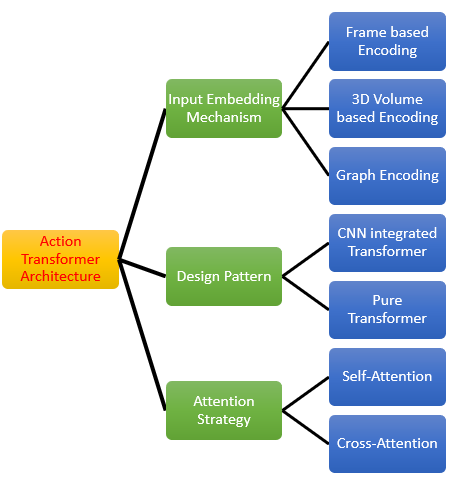} 
\caption{Taxonomy 1: Action Transformer classification based on architectural approaches. This taxonomy classifies action transformers according to their architecture, which includes their backbone structure, input encoding mechanism, and attention strategy.}
  \label{fig:T1}
\end{figure}

This section comprehensively reviews existing visual Transformers for action recognition and groups them into different categories according to appropriate criteria. 
We start with the classification of action transformers according to their architecture, which includes their backbone structure, input encoding mechanism, and attention strategy. Such a classification is shown in Fig.~\ref{fig:T1}. It also includes a discussion about CNN-based Action Transformer methods that utilize Transformer to enhance the representation learning of CNNs. Due to the negligence of local information in the original transformer backbone, the CNN Enhanced Transformer employs an appropriate convolutional inductive bias to augment the visual Transformer, while the Local Attention Enhanced Transformer redesigns patch partition and attention blocks to improve their locality. On the other hand, Transformer enhanced CNN design for action recognition has also been discussed. Finally, pure transformer-based action transformers are presented.

In the second type of classification, we use different criteria of modality to classify different approaches for action transformers. It includes unimodal and multimodal approaches that are further divided concerning the presented modality.  Finally, we group action transformers into four different clusters based on their intended usability and final task such as action classification, detection, segmentation, and anticipation. 

For each category, we present an overview, a brief discussion of some representative approaches, their pros and cons, and a  table to summarize representative work.

\subsection{Taxonomy 1: Classification based on Network Architecture }

In this section, we discuss and classify different action transformer approaches based on three criteria that include backbone architecture, encoding scheme, and attention mechanisms. We also discuss some representative approaches in detail to illustrate the concepts. 

Our first criterion is backbone network architecture and pattern, which are vital to classify different approaches. Based on existing approaches from our literature survey, we classify all action transformers into two categories: CNN integrated transformer-based action recognition and pure transformer-based architectures.   

\textbf{CNN- Integrated Designs}:  The first category is CNN integrated transformers. CNN \cite{CNN} and RNN \cite{RNN}  have been used in the past to represent and solve sequential problems. During the last decade, a plethora of deep learning approaches for action recognition have been proposed predominantly based on CNN architectures and their 3D extensions. These include action recognition from different modalities and applications on different challenging action recognition benchmarks. CNN features, from both 2D and 3D convolutional networks, have been integrated into transformer-based designs to recognize action categories. The major step in such approaches is the extraction of representative features that can be extracted with a pre-trained CNN or learnend in an end-to-end fashion. A successful scheme is based on the extension of Convolutional Neural Networks to 3D that automatically learns spatio-temporal features. Such methods can employ a traditional classifier or an end-to-end network that includes fully connected layers and softmax for action classification. An alternative approach is to use sequential networks like  Recurrent Neural Network or LSTM \cite{LSTM} and trained it to classify each sequence considering the temporal evolution of the learned features for each timestep. Table 2 provides some salient techniques based on this category.  

Below, we discuss only the most relevant works based on high citations or architectural relevance to this category.

\begin{itemize}

 \item One of the earlier works that used the term Action Transformer is Video action Transformer Network \cite{VATN}. It is a  CNN integrated  Transformer architecture that aggregates features from the spatiotemporal context around the person whose actions it tries to classify. It employs person-specific, class-agnostic queries to spontaneously learn to track individual people and to pick up on semantic context from the actions of others. Its attention mechanism learns to emphasize hands and faces, which are often crucial to discriminate an action.  This model was trained and tested on the Atomic Visual Actions (AVA) dataset, outperforming the state-of-the-art by a significant margin, and still is one of the highly cited papers that uses a transformer for action recognition.  Fig.~\ref{fig:ATN} provides an architectural overview of the approach as presented by the authors.  

 \item  The second model based on 3D CNN integrated transformer is ConvTransformer Network \cite{COVER} for action detection that comprises three main components: (1) a Temporal Encoder module that explores global and local temporal relations at multiple temporal resolutions, (2) a Temporal Scale Mixer module that effectively fuses multi-scale features, creating a unified feature representation, and (3) a Classification module which learns a center-relative position of each action instance in time and predicts frame-level classification scores. Consider features of video segments extracted by a 3D CNN as inputs to MS-TCT, which embed spatial information latently as channels. Specifically, it uses an I3D backbone to encode videos. Each video is divided into T non-overlapping segments (during training), each of which consists of 8 frames. Such RGB frames are fed as an input segment to the I3D network. Each segment-level feature (output of I3D) can be seen as a transformer token of a time-step (i.e., temporal token). It stacks the tokens along the temporal axis to form a video token representation, to be fed into the Temporal Encoder.

 \item Another representative work in this category is  Spatio-Temporal Attention Network (STAN) \cite{STAN} which uses a two-stream transformer architecture to model dependencies between static image features and temporal contextual features. It gains data and computational efficiency by introducing inductive bias via convolution. This approach replaces the tokenization method with both image spatial and context temporal scene features extracted from pre-trained convolutional neural networks (CNNs). It then uses a two-stream transformer architecture to model temporal dependencies between the scene features for classifying long videos of up to two minutes in length. 

  \end{itemize}
 
\begin{table*}
\begin{center}
 \begin{tabular}{|l|l|l|p{4cm}|}
\hline
\multicolumn{4}{ |c| }{{\cellcolor{green}} Representative Methodologies} \\
\hline
\hline
 Model Acronyms & Citation & Feature backbone & Attention \\ [0.5ex] 
 \hline\hline
VTCE & \cite{VTCE} &
  CNN-ResNET-50 & 
   Multi-headed self-attention
         \\
\hline
FRAB & \cite{FRAB}&
  ResNeXt-101  &
    Multi-headed self-attention\\
\hline
TadTR  & \cite{TadTR} &
  I3D,TSP &
    Temporal deformable self and cross attention\\
\hline
CBT & \cite{CBT}&
  S3D, &
   Multi-headed self-attention \\
\hline
STAN & \cite{STAN}&
  ResNet18  &
   Multi-headed self-attention \\
\hline
MS-TCT & \cite{MS-TCT}&
  I3D  &
   Multi-head self-attention \\
\hline
ST-TR & \cite{ST-TR}&
  GCN and 2D convolutions &
   Temporal Self-Attention \\
   \hline
VTN & \cite{VTN}&
  SE-ResNeXt-101 &
   Scaled Multiplicative Attention \\
\hline
TPT &  \cite{TPT}&
  I3D  &
  Clip-level self-attention \\
\hline
VideoLightFormer & \cite{VideoLightFormer}&
  2D CNN Backbone &
    Multi-headed self-attention \\
\hline
Actor-Transformer & \cite{Actor-Transformer}&
  I3D and HR Net &
     Multi-headed self-attention \\
\hline
UGPT & \cite{UGPT}&
  3D ResNet-101  &
  Probabilistic Attention \\
\hline
VATN & \cite{VATN}&
  I3D and RPN &
  Multi-headed self-attention \\
\hline
KA-AGTN & \cite{KA-AGTN}&
  GCN  &
    Temporal Kernel Attention \\
\hline
RTDNet & \cite{RTDNet}&
  I3D and GCN &
  Multi-headed self-attention  \\
\hline
STAT & \cite{STAT}&
  3D CNN &
  Shrinking Temporal Attention \\ \hline
\end{tabular}
\captionsetup{singlelinecheck=false, font=small, labelfont=bf}
   \caption{CNN-Integrated Approaches: These tables list the most recent approaches related to this category by mentioning their model acronym, citation, CNN features network backbone, and transformer-based attention mechanism. }
  \end{center}
    \end{table*}


The major  strengths and weaknesses of CNN integrated transformer approaches are as follows:

\textbf{Strengths: }
 \begin{itemize}
 
\item These methods are based on trusted and proven CNN 2D or 3D features that have demonstrated their strengths in extracting meaningful spatial and temporal information required to recognize human actions.

\item Most of these visual features are supported with explainable visualization to show their strengths in extracting required information that is essential in action recognition processes. Convolution has proved to be a valuable process to learn visual features hierarchically.  

\item Most of the CNN-based models are well established, easy to replicate their designs and pre-trained models are available that make it feasible to integrate them into any design. Even attention mechanisms can be integrated into these approaches. 
  \end{itemize}

\textbf{Weaknesses:}

\begin{itemize}
\item	Although CNN models are translation invariant, they face a hard time dealing with positional and temporal information in videos and if the object of interest (actor) has variations in rotations and scaling. This weakness of CNN affects the transformer later on in the recognition process.

\item	During the striding process, the CNN completely loses all the information about the composition and position of the components and they transmit the information further to a neuron that might not be able to classify the right label for actions. Data augmentation usually revolves around mitigating such issues by flipping or rotating the training data in small amounts.  However, data augmentation does not solve the worst-case scenario as real-life situations have complex pixel manipulation in activity recognition. This is the same reason why CNN-based approaches are vulnerable to adversarial attacks. 

\item	A convolutional neural network is significantly slower due to an operation such as max pool and requires a large dataset to process and train the neural network. However, this applies to transformer models. 

\end{itemize}


\textbf{Pure Transformer Designs}: In pure transformer methods, convolutions are abolished and transformers are the sole building block of the architecture. This category is especially important for researchers in the area to know the real strength of transformers to encode and perform complex computer vision tasks such as action recognition. 

\begin{figure}
\centering
\includegraphics[width=160 mm]{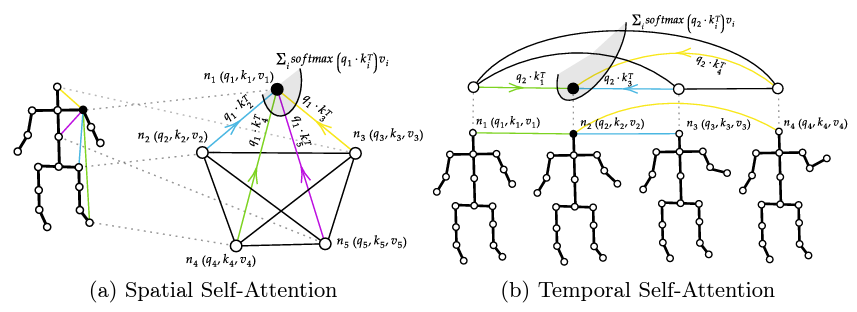} 
\caption{An overview of Spatial Self-Attention (SSA) and Temporal Self-Attention (TSA) in the skeleton-based transformer-based approach.Image adapted from \cite{ST-TR}. }
\label{fig:Skeleton}
\end{figure}

There is growing interest in using pure transformer based models, especially extending the video transformer concept for action and activity recognition and various approaches are now available in the literature that follow this architecture by providing and discussing various encoding techniques and attention mechanisms. Most of the approaches in this category follow a vision transformer-based approach that considers different encoding schemes to tokenize videos. Encoding and embedding include 2D frame-based approaches and 3D volume-based approaches. Similarly, the third branch is graph-based encoding, which is especially used for skeleton action recognition tasks. Fig.~\ref{fig:Skeleton} provides an architectural overview of one of the skeleton based approaches for action recognition.  It is interesting to know if transformers can replace or outperform convolution-based approaches in performing such complex vision tasks.  Table 3 provides some important approaches that have appeared recently based on this architectural approach. 

Here, we discuss only the most relevant work based on citations and architectural relevance to this category.

\begin{figure}
\centering
\includegraphics[width=120 mm]{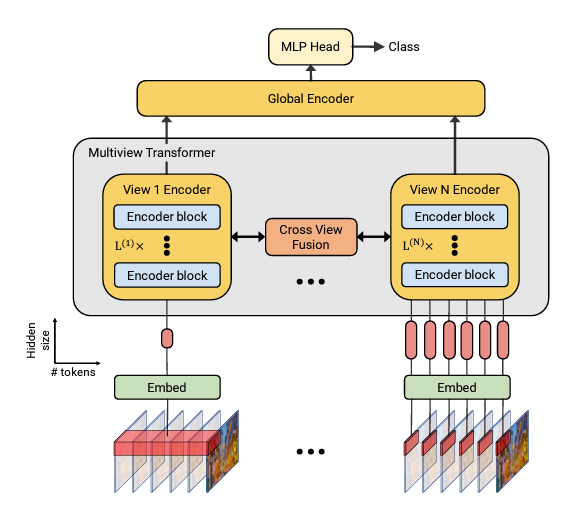} 
\caption{Overview of Multiview Transformer, a pure transformer-based approach for action recognition. Image adapted from \cite{MTV}. }
\label{fig:MTV}
\end{figure}

\begin{itemize}
\item One of the earliest pure visions transformer-based approaches is Timesformer \cite{Timeformer} that adapts the standard Transformer architecture to videos by enabling spatio-temporal feature learning directly from a sequence of frame-level patches. It is a fully convolution-free approach to video classification built exclusively on self-attention in space and time. It compares different self-attention schemes and suggests that ``divided attention'', where temporal attention and spatial attention are separately applied within each block, leads to the best video classification accuracy among the design choices considered. Timesformer takes as input a clip consisting of $F$ RGB frames of size $H$, and $W$ sampled from the original video. It then decomposes each frame into $N$ non-overlapping patches, each of size $P$, such that the $N$ patches span the entire frame. It then linearly maps each patch into an embedding vector using a learnable matrix. The Transformer consists of $L$ encoding blocks. At each block, a query/key/value vector is computed for each patch from the representation encoded by the preceding bloc. Self-attention weights are computed via dot-product. Then, the concatenation of these vectors from all heads is projected and passed through an MLP, using residual connections after each operation. The final clip embedding is obtained from the final block for the classification token.

\begin{figure}
\centering
\includegraphics[width=160mm]{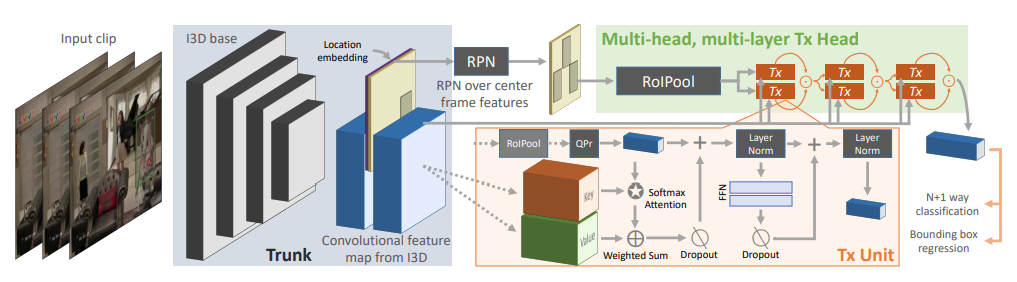} 
\caption{A highly cited video Action Transformer Model  for video based action recognition. Image adapted from \cite{Actionformer}. It is one of the earlier works that used the term Action Transformer and uses a CNN integrated  Transformer architecture that aggregates features from the spatiotemporal context around the person whose action is being classified.}
\label{fig:ATN}
\end{figure}

\item  The second most important work in this category is Video Transformer (VidTr)\cite{Vidtr} with separable attention for video classification is a stacked attention-based architecture for video action recognition. Compared with commonly used 3D networks, VidTr can aggregate spatiotemporal information via stacked attentions and provide better performance with higher efficiency. It introduces multi-head separable attention (MSA) by decoupling the 3D self-attention to spatial attention and temporal attention. It also uses temporal downsampling to remove redundancy in video data representation. 

\item One of the top performing methods on various action recognition benchmark dataset leaderboards is Multiview Transformers for Video Recognition (MTV) \cite{MTV}. Separate encoders are used to represent each perspective in the input video, and lateral connections are used to combine data from multiple perspectives into a single coherent representation. Additionally, it only calculates self-attention among tokens extracted from the same temporal index within each transformer layer. The model's computational burden is drastically diminished as a result. It is also believed that the multiview encoder's fusion of information from other views and the global encoder's aggregation of tokens from all streams render self-attention along all spatiotemporal tokens unnecessary.  It uses cross-view attention (CVA), a straightforward method of combining information between different views by performing self-attention jointly on all tokens. Another method used is transferring information between tokens from two views,   by an intermediate set of $B$ bottleneck tokens. It sequentially fuses information between all pairs of two adjacent views. Finally, it aggregates the tokens from each of the views with the final global encoder. Fig.~\ref{fig:MTV} provides an architectural overview of the approach as presented by the authors.  

  \end{itemize}

\begin{table*}
\begin{center}
\begin{tabular}{|l|l|l|p{5cm}|}
\hline
\multicolumn{4}{ |c| }{{\cellcolor{green}} Representative Methodologies} \\
\hline
\hline
 Model Acronyms & Citation & Encoding Methodology & Attention \\ [0.5ex] 
 \hline\hline
 STPT &  \cite{STPT} &  
  Patch Conditional  Encoding &
  Spatio-Temporal Attention \\
\hline
ASFormer & \cite{ASFormer} &
  Frame base Patches &
 Spatio-Temporal Attention \\
\hline
AcT &\cite{AcT} &
  Frame base Patches &
Spatio-Temporal Attention \\
\hline
  DVT  & \cite{DVT}&
  Frame base Patches &
  Deformable S-T Attention \\
  \hline
  ViVIT  & \cite{vivit} &
  Tubelet Embedding &
  Factorised dot-product Attention \\
\hline
Video-Swin &\cite{V-SWIN} &
  3D patch partitioning &
  MHA non-overlapping 3D windows \\
\hline
RViT  &\cite{RViT}&
  Frame Patch Embedding  &
  Attention Gates \\
\hline
MTV & \cite{MTV}&
  Multiview tokenization &
  Cross-view attention \\
\hline
DirecFormer & \cite{DirecFormer} &
 Frame Patch Embedding  &
  Directed Temporal Attention\\
\hline
Timesformer & \cite{Timeformer} &
  Frame Patch Embedding  &
  Divided Space-Time  Attention \\
\hline
STAR &\cite{STAR}&
  Segmented Sequential Encoding,
 &   Segmented linear attention
 \\ \hline
\end{tabular}
 \captionsetup{singlelinecheck=false, font=small, labelfont=bf}
   \caption{Pure Transformer-based Approaches: This table lists the most recent approaches related to this category by mentioning their model acronym, citation, encoding methodology, and transformer-based attention mechanism. }
  \end{center}
    \end{table*}


The major  strengths and weaknesses of the Pure transformer approaches are as follows:

\textbf{Strengths: }
 \begin{itemize}
 
 \item  Vision Transformers are rapidly starting to dominate many applications in Computer Vision and it also applied to action recognition tasks as such models are top performers on major action recognition datasets. On various leaderboards, they have achieved higher accuracies on large data sets due to their higher modeling capacity and lower inductive biases as well as their global receptive fields. 
 
 \item These methods have a pure attention-based approach and reply on attention maps rather than feature maps. It helps to reduce the redundancy of irrelevant content and focus on relevant content. Therefore in these models, visual features are dynamically re-weighted based on the context. They mostly use a higher number of parameters that enables them to grasp the complexity of the data in challenging tasks such as action recognition. 
 
\item  Pure Vision transformers better encode positional and temporal information that is vital for action recognition. In addition, they are proven to be robust against naturally corrupted patches.  The transformer can be applied to model relationships between various basic visual elements, making them suitable for action recognition. That is one of their application in pose or skeleton-based action recognition tasks. 

\item  Pure Vision transformer is a newly adapted framework compared to CNN that took time to evolve and fix its limitations and improve its performance. Transformers have more potential to evolve as research in this direction has just started recently.  

 \end{itemize}

\textbf{Weaknesses: }
 \begin{itemize}
\item Convolutions with shared weights explicitly encode how specific identical patterns are repeated in images. This inductive bias ensures easy training convergence.  Vision Transformers-based designs do not enforce such strict inductive biases. This makes them harder to train.
 \item Even though pure transformers have shown State-of-the-Art (SOTA) performance on many action recognition datasets, they do not necessarily outperform CNNs across the board, especially on lightweight action recognition tasks \cite{Evaluating}. Their implementation on edge devices is especially a challenge. 
\item CNN-based methods have proved their strength and effectiveness over time in various scenarios and challenges. Vision transformer-based models have yet to prove their effectiveness in terms of efficiency for wide adaptation. 
  \end{itemize}


\textbf{Encoding and Embedding Schemes}: The first step to applying transformer on video data is to find a representation that could be used reasonably in their architecture. There are two challenges to using this architecture, one is its computational expense and the other is its sequential nature. 

The input to the transformer is a 1D embedding. Input 2D images are processed by first flattening them into a series of image patches of the same size. The velocity data can be put to good use by a graph-aware transformer (GAT) \cite{AGT}, which can then use that data to infer discriminative spatial-temporal motion features from the skeleton graphs' sequence.
The theoretical advantage should be that the network should be able to grasp the pattern from the encoding and thus generalize better for longer sentences. With one-hot position encoding, embeddings of earlier positions can be learned much more reliably than embeddings of later positions.

Therefore, another categorization could be based on various encoding and embedding schemes for transformer-based approaches. These approaches vary from CNN feature encoding, frame patch encoding, 3D tubes, graph convolutions, positional encoding, actor level feature encoding, 3D patch partitioning, segmented sequence embedding, and conditional positional encoding.

1- \textbf{Frame based Patch encoding and linear embedding}: For instance, Timesformer accepts as input a clip with $F$ RGB frames of size $H$ and $W$ taken from the original video as samples. Each frame is then dissected into $N$ overlapping patches of size $P$, with each patch covering the entire frame. Then, a learnable matrix is used to linearly map each patch onto an embedding vector. There are $L$ encoding blocks inside the Transformer. Each patch has its positional data superimposed on top of a linear map into an embedding. When fed into a Transformer encoder, the resulting vector sequence can be understood as token embeddings.

A deformable video transformer \cite{DVT} takes as input video clips consisting of $T$ RGB frames sampled from the original video. The video is converted into a sequence of S·T tokens. The tokens are obtained by decomposing each frame into Spatches which are then projected to D-Dimensional space through a learnable linear transformation. This tokenization can be implemented by linearly mapping the RGB patches of each frame or by projecting space-time cubes of adjacent frames. Separate learnable positional encoding is then applied to the patch embeddings for the spatial and temporal dimensions.

2- \textbf{3D volume encoding and embedding}: ViViT \cite{vivit} linearly projects the input volume after extracting non-overlapping Spatio-temporal "tubelets''. This technique is analogous to a 3D convolution and represents a 3D extension of ViT's embedding. Tokens are drawn from the time, height, and width dimensions of a tubelet. Because of this, the computation time grows as the tubelet size decreases. Spatio-temporal information is fused during tokenization in this approach, as opposed to "Uniform frame sampling," where only temporal information is fused across frames by the transformer.

A video Swin Transformer \cite{V-SWIN}, receives an input video, with a total of T frames. A 3D patch is considered a token in Video Swin-Transformer. 3D tokens obtained from the 3D patch partitioning layer are each composed of a 96-dimensional feature. Next, a linear embedding layer is used to map the token features to the dimension C. It extended SWIN-Transformer model to video domain. 

3- Graph-based embeddings: For skeleton-based action recognition, the representation of actions consists of two core parts: (1) the adjacency matrix representing the structure of the human skeleton graph. (2) The learned skeleton tensor representing the spatiotemporal actions, which is the set of feature vectors corresponding to the vertex in the frame and containing channels. Good skeleton graph embeddings accurately and dynamically reflect the spatiotemporal dependencies between joints over time. The interaction of joints is a key attribute of action. 

SGT\cite{SGT} uses an adaptive graph strategy where the input contains a skeleton tensor and an adjacency matrix. The output is an adaptive skeleton tensor. The adaptive adjacency matrix acts on the original skeleton tensor through the embedding function (2D convolution is selected in the implementation).

Spatial-Temporal Transformer (ST-TR) \cite{ST-TR}, uses in-frame self-attention to compute independent correlations between every pair of joints in each frame in order to extract low-level features embedding the relations between body parts. After applying trainable linear transformations to the node features, which are shared across all nodes of parameters, a query vector, key vector, and value vector are first computed for each node of the skeleton given the frame at time t. Afterwards, a query-key dot product is used to calculate a weight for each set of body nodes, where the weights represent the degree to which the nodes are correlated with one another. A new embedding for the node is then calculated using a weighted sum based on the score applied to each joint value.

\textbf{Attention Schemes}: We discuss and classify different Self-Attention and Cross Attention schemes for transformer based action recognition. One building block of machine learning transformers is attention, and more specifically, self-attention. It is a computational primitive that helps a network discover the  contextual relevance and alignments hidden in the input data by quantifying the interactions between pairs of entities. Focused mental energy has been shown to be an essential ingredient for vision networks to become more robust. Multiple parallel computations are performed by the Transformer's Attention module. These are generally referred to as Attention Head. The Query, Key, and Value are separated into $N$ equal parts by the Attention module, and then each of those parts is sent to its own Head.

Self Attention, also known as intra -Attention, is an attention mechanism that computes a representation of a sequence by relating its various positions. It has been successfully used for automated reading, abstract summarization, and making image descriptions. The Transformer architecture includes a cross-attention mechanism that utilises a combination of two distinct embedding sequences. It is necessary that the lengths of both sequences be equal. Different modes can be used in each of the two sequences (e.g. text, image, sound).

Different attention strategies for action transformers include Multi-headed self-attention, Cross-Modality Attention, Temporal deformable attention, Temporal deformable cross-attention, Scaled Multiplicative attention, Temporal Kernel Attention, Shrinking Temporal Attention,  Deformable Space-Time Attention,  Factorised dot-product attention, Directed Temporal Attention, Sparse Local Global  Attention. Axial Attention, Joint Space-Time Attention, Divided Space-Time Divided Space-Time  Attention, and Segmented linear attention.

\begin{figure}
\centering
\includegraphics[width=165 mm]{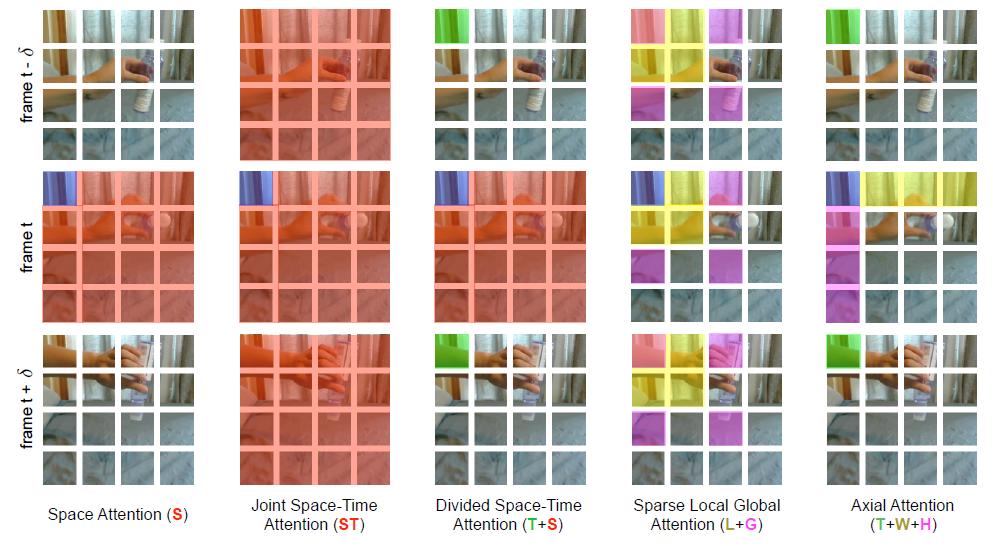} 
\caption{Visualization of the five space-time self-attention schemes as described by the Timesforer Model. Image adapted from \cite{Timeformer}. }
\label{fig:SA}
\end{figure}

\textbf{\textit{Self-attention via dot-product}}:  Five space-time self-attention schemes are discussed in Timesformer. Each video is broken down into a series of $16\times16$  pixel patches at the frame level. These attention mechanisms are shown in Fig.~\ref{fig:SA} . For each scheme, the blue dot represents the query area, while the other colours designate the space-time region of self-attention that surrounds the query area. A blue patch's level of focus does not take into account a nearby colourless patch. Coordinated colour usage suggests thinking about more than one spatial or temporal dimension or more than one set of neighbours (e.g., "T+S" or "L+G"). Remember that self-attention is computed for each patch in the video clip, with each patch acting as a query. Furthermore, we see that the clip as a whole, not just the two highlighted frames, follows this pattern of attention.

\textit{\textbf{Factorised self-attention}} introduced in ViViT, in contrast, computes multi-headed self-attention across all pairs of tokens, at layer , it factorizes the operation to first only compute self-attention spatially (among all tokens extracted from the same temporal index), and then temporally (among all tokens extracted from the same spatial index). Each self-attention block in the transformer thus models Spatio-temporal interactions but does so more efficiently than Model 1 by factorizing the operation over two smaller sets of elements. 

Multiview video transformer employs \textbf{\textit{cross-view attentio}}n (CVA) to pool data from multiple perspectives into a single self-attention operation for all tokens. Given the computational impossibility of self-attention in video models, this method sequentially fuses information between all pairs of two adjacent views, where the views are ordered in terms of increasing numbers of tokens (i.e., token density). It does this by first projecting the keys and values to the same dimension, as the hidden dimensions of the tokens between the two views can be different.

TadTR \cite{TadTR} uses a temporal deformable cross-attention (TDCA) module. The motivation here is that multiple actions in one video are often related. For example, a cricket shot action often appears after a cricket bowling action. To make an action prediction, each query extracts relevant context information from the video via the TDCA module.

Multimodal cross-attention is used in MM-ViT. MM-ViT \cite{MM-ViT} operates on a compressed video clip $V$. The vision modalities consist of T sampled I-frames, motion vectors, and residuals of height H and width W. MM-ViTI-Joint Space-Time-Modality Attention: Similar in spirit to the “Joint Space-Time Attention” in Timesformer and the “Spatio-Temporal Attention” in ViViT, each transformer layer of this model measures pairwise interactions between all input tokens. Concretely, MM-ViT I consists of L transformer layers.

\subsection{Taxonomy 2: Action Transformers Modality}

\begin{figure}
\centering
\includegraphics[width=140 mm]{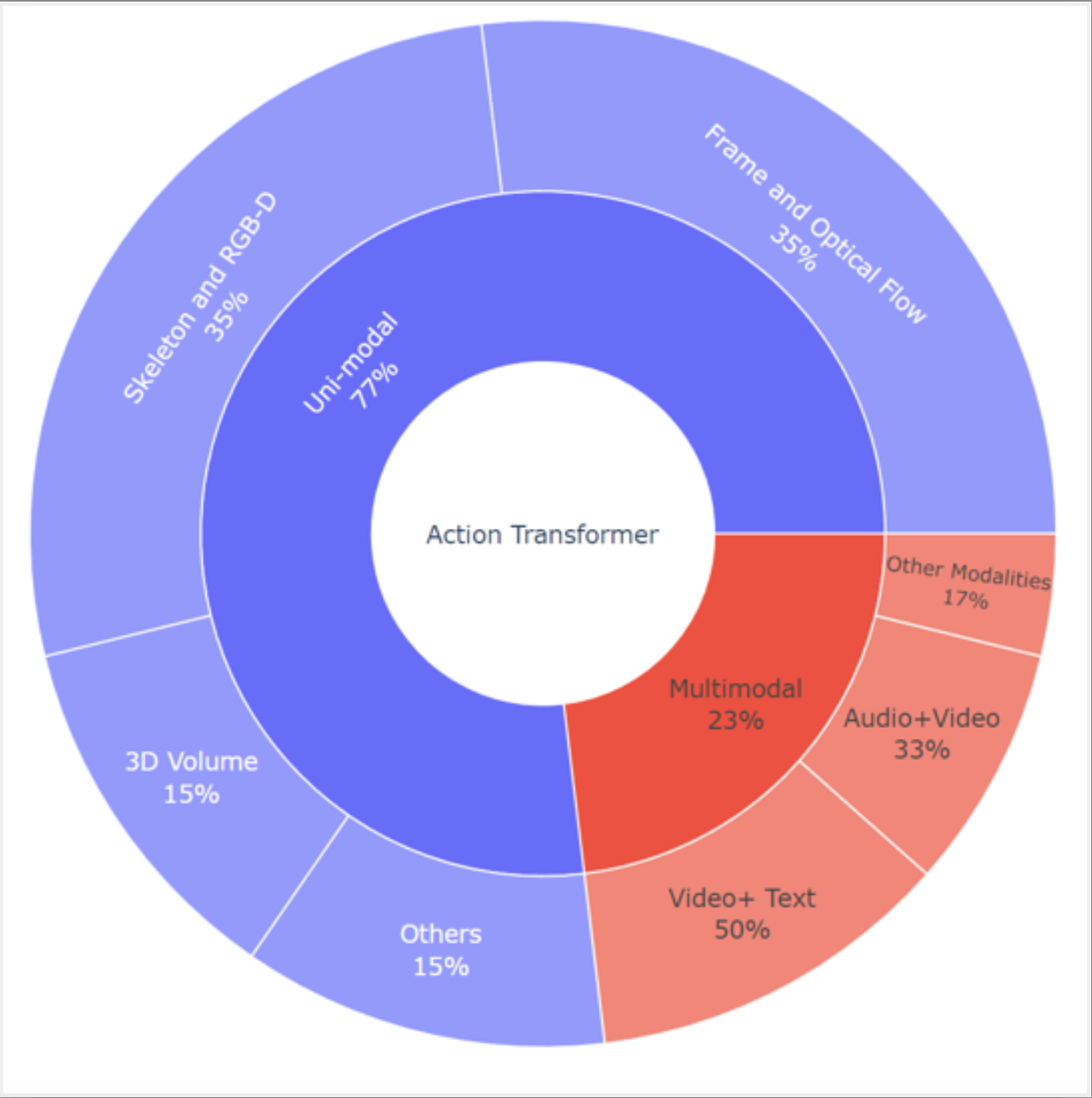} 
\caption{Taxonomy 2: Action Transformer categorization By Modality.It shows the compatibility of Transformers with multiple modalities for action recognition tasks. majority of the methods are developed for single modalities. However, multimodal transformers are getting popularity due to the rise in the number of multimodal approaches. }
\label{fig:T2}
\end{figure}

The compatibility of Transformers with multiple modalities in both discriminative and generative tasks has been the subject of extensive research as of relatively recent times. Visual and linguistic modalities, for example, are typically associated with particular sensors that facilitate distinct modes of interaction. Fig.~\ref{fig:T2} provides a visualization of classification of vision transformer based action recognition approaches categorized by modality.

The Transformer family can be thought of as a special case of general graph neural networks, despite its generality. Self-attention, in particular, can treat all inputs as if they were nodes in a single graph by focusing on the global (not local) patterns. Since the embedding of each token can be thought of as a node in a graph, this inherent trait ensures that Transformers can function in a modality-agnostic pipeline that is flexible enough to accommodate a wide range of modalities.

Self-attention and its variants play a central role in processing cross-modal interactions (e.g., fusion, alignment) in multimodal Transformers. Two-stream Transformers can experience cross-modal interactions if Q (Query) embeddings are swapped or exchanged in a cross-stream fashion. Cross-attention, also known as co-attention, describes this strategy. Cross-attention pays attention to each modality dependent on the other without increasing computational complexity, but if considered for each modality, this approach fails to perform cross-modal attention globally and thus loses the entire context. Since there is no self-attention to the self-context within each modality, learning cross-modal interaction requires two-stream cross-attention.

A wide variety of data modalities, including RGB, skeleton, depth, infrared, point cloud, event stream, audio, acceleration, radar, and WiFi signal, can be used to represent human actions. These modalities encode different sources of useful yet distinct information and offer different benefits depending on the application scenarios. For this reason, many prior works have attempted to probe various strategies for HAR employing various modalities.

Here, we discuss different modalities that are used to recognize actions using transformer architecture. Table 3 describes some important modalities regarding action transformers.

Images and videos (sequences of images) captured with RGB cameras that attempt to replicate what human eyes see are often referred to as being in the RGB modality. The captured scene's RGB data can be easily retrieved and contains extensive appearance details. Visual surveillance, autonomous navigation, and sports analysis are just a few of the many uses for HAR that is based on the RGB colour space \cite{Realistic,Drivering,surveillance}.

Joint trajectories, which are indicative of meaningful human motion, are encoded in skeletal sequences. This means that skeletal information is another valid data type for HAR. Applying pose estimation algorithms on RGB videos \cite{Pose2}  or depth maps \cite{posep}.  will yield the skeleton data. Additionally, motion capture systems can be used to collect this data. Human pose estimation is notoriously sensitive to changes in camera angle. However, reliable skeleton data can be obtained from motion capture systems that are not view or lighting sensitive. However, motion capture systems are inconvenient to implement in a wide variety of practical contexts. Because of this, many recent skeleton-based HAR works have relied on skeleton data derived from depth maps or RGB videos.

The Contrastive Captioner (CoCa) \cite{CoCa}, is a minimalist multimodal design that combines the capabilities of contrastive and generative methods by pre-training an image-text encoder-decoder foundation model with contrastive loss and captioning loss. While conventional encoder-decoder transformers have all decoder layers pay attention to encoder outputs, CoCa only uses the first half of its decoder layers to encode unimodal text representations before cascading the remaining decoder layers, which cross-attend to the image encoder, to create multimodal image-text representations. It employs a captioning loss on the multimodal decoder outputs, which makes autoregressive predictions of text tokens, and a contrastive loss between unimodal image and text embeddings. Due to the two training goals using the same computational graph, we can achieve high computation efficiency with low overhead. Coca is trained in its entirety, from scratch, on web-scale alt-text data and annotated images, with all labels being treated as text. This unifies natural language supervision for representation learning. From an empirical perspective, we see that CoCa is able to outperform the state-of-the-art on a wide variety of visual recognition downstream tasks with either zero-shot transfer or minimal task-specific adaptation.

MERLOTReserve \cite{Merlot} is a model that represents videos jointly over time–through a new training objective that learns from audio, subtitles, and video frames. Given a video, it replaces snippets of text and audio with a MASK token; the model learns by choosing the correct masked-out snippet. Its objective learns faster than alternatives, and performs well at scale: it is pre-trained on 20 million YouTube videos.

\begin{table*}
\begin{center}
\begin{tabular}{|l|l|l|p{3cm}|p{4cm}|}
\hline
\multicolumn{5}{ |c| }{{\cellcolor{green}} Representative Methodologies} \\
\hline
\hline
 Model Acronyms & Citation & Modalities & Architecture & Datasets Used \\ [0.5ex] 
 \hline\hline
 AAA & \cite{AAA} &
  Video and Audio &
  Pure-Transformer &
  EPIC-Kitchens, CharadesEgo   \\
 \hline
EAO  &\cite{EAO}&
  Audio, Text, and Video &
  Pure-Transformer &
  HowTo100M \\
  \hline
COVER & \cite{COVER}&
  image  and Video co-training &
  Pure-Transformer &
  Kinetics-400, 600, 700, SSV2 and MIT \\
  \hline
CCD & \cite{CCD} &
  Text  and Video &
  CNN integrated Transformer &
  Tasty video dataset and Recipe1M \\
  \hline
ActionCLIP  & \cite{Actionclip} &
  Text and Video &
  Pure-Transformer &
  Kinetics-400 \\
  \hline
LST & \cite{LST}&
  Text and  Skeleton &
  Graph Convolution-based Transformer &
  NTU RGB+D and  NTU RGB+D 120 \\
  \hline
Speech2Action & \cite{speech2action} &
  Speech and Video &
  CNN Integrated Transformer &
  IMSDb dataset \\
  \hline
MM Mix & \cite{MM-Mix} &
  spectrogram, optical flow &
  Pure Transformer &
  Epic-Kitchens \\
  \hline
LOSOCV & \cite{LOSOCV}&
  Skeletal and acceleration  &
  Pure Transformer &
  The NCRC dataset \\
  \hline
VATT & \cite{Vatt} &
  Audio, Text, and Video &
  Pure Transformer &
  Kinetics-400, 600,700, and  MIT \\
  \hline
EAMAT & \cite{EAMAT} &
  Text and  Video &
CNN integrated Transformer &
  Charades-STA and TACoS datasets \\
  \hline
MTCN  &\cite{MTCN}&
  Speech and Video &
  Pure Transformer &
  EPIC-KITCHENS-100 \\
  \hline
MM-ViT & \cite{MM-ViT}&
  Appearance, motion, and Audio &
  Pure-Transformer &
  UCF-101,SSv2, Kinetics-600 \\
   \hline
UDAVT &\cite{UDAVT}&
  Different video Modalities &
  Pure-Transformer&
  HMDB to UCF, Kinetics to NEC-Drone 
 \\ \hline
\end{tabular}
 \captionsetup{singlelinecheck=false, font=small, labelfont=bf}
   \caption{Taxonomy2: Modality-based Transformer Approaches: This table lists the most recent approaches related to this category by mentioning their model acronym, citation, modality, architecture, and datasets. To note, SSV2 is Something-something-v2 and MIT is Moments in time abbreviation.  }
  \end{center}
    \end{table*}


\subsection{Taxonomy 3: Action Transformer End-task}

Human behavior is an important aspect of how we operate in our world, so  computer vision models need to learn to detect and distinguish human actions as an end task. 

Time is fundamental to existence, and as such, all of life can be thought of as a series. While a standard neural network could be used to perform machine learning on sequential data (such as text, speech, video, etc.), doing so would be constrained by the network's fixed input size. Action recognition models are usually trained using large image or video databases.  Action recognition can be broken down into the following tasks:

\begin{table*}
\begin{center}
\begin{tabular}{|l|l|l|l|}
\hline
\multicolumn{4}{ |c| }{{\cellcolor{green}} Representative Methodologies} \\
\hline
End-task & Model & Datasets Used \\ \hline
\multirow{5}{*}{Action Segmentation} 
& Lin. Attn. Transf.&\cite{Lin.Attn.Transf.} & Breakfast , Hollywood Extended, 50 Salads \\
& U-Transformer & \cite{U-Transformer}    & 50Salads,GTEA,Breakfast          \\
& CETNet &\cite{CETNet}& 50Salads, Georgia Tech Egocentric Activities, Breakfast \\
& nRM  &\cite{nRM}  & PKU Multi-Modality Dataset - PKU-MMD        \\
& SCT  & \cite{SCT}  & Breakfast,Hollywood Extended,MPII 2 Cooking,   \\ \hline
\multirow{5}{*}{Action Localization} 
& AGT  &\cite{AGT}            & THUMOS14+ Charades+ EPIC-Kitchenes-100      \\
& ActionFormer  &\cite{Actionformer}   & THUMOS14 +ActivityNet 1.3 +EPIC-Kitchens 100 \\
& TAL Framework &\cite{TALFramework}  & CVPR2021 HACS Challenge \\
&LSTM-T&\cite{LSTM} & Charades-STA and TACoS   \\
&Ag-Trans & \cite{Ag-Trans}  & THUMOS14+ ACTIVITYNET1.3+MUSES    
  \\ \hline
 \multirow{5}{*}{Action Detection} 
&TubeR & \cite{TubeR}   & AVA, UCF101-24 and JHMDB51-21.                 \\
&MS-TCT &\cite{MS-TCT}   & Charades, TSU and MultiTHUMOS                  \\
&OADT  &\cite{Oadtr}    & Epic-KITCHENS-100                              \\
&Stargazer &\cite{Stargazer} & Runners-up in the AI City Challenge 2022 \\
&LSTR  &\cite{LST}    & THUMOS’14, TVSeries, and HACS Segment          \\
&OadTR &  \cite{Oadtr}  & HDD, TVSeries, and THUMOS14    
 \\ \hline
 
  \multirow{5}{*}{Action Anticipation} 
&HORST & \cite{HORST}   & Something-Something  and EPIC-Kitchens    \\
&CCD  &\cite{CCD}     & Tasty video dataset,Recipe1M dataset.                                                 \\
&URM-CTP& \cite{URM-CTP}  & EPIC-Kitchens-55, EPIC-Kitchens-100  \\
&MM-Trans.&\cite{MM-Transformer} & 50 Salads, Breakfast,  EPIC-KITCHENS55                                        \\
&AVT &\cite{AVT}& EpicKitchens-100,EGTEA Gaze+, 50-Salads \\
&FUTR& \cite{FUTR}     & Breakfast, 50 Salads,        
 \\ \hline
 
\end{tabular}
\captionsetup{singlelinecheck=false, font=small, labelfont=bf}
   \caption{ Taxonomy 3: This classification of transformer-based action recognition methods is based on their end task which includes action detection, localization, segmentation, and anticipation. To note, this table did not list action classification approaches due to the majority of approaches belonging to this category. }

  \end{center}
    \end{table*}


Images or videos are analysed in an effort to determine what kind of action they depict, such as "cooking," "writing," etc. Machine learning models that can recognise human actions in video are called \textbf{"action classifiers."} The term "jumping jacks" can be predicted, for instance, by an action classifier that has been trained to recognise exercise movements by observing a video of a person performing the activity.

\textbf{Action Segmentation}, in its most basic form, seeks to time-slice an untrimmed video into discrete segments and label each segment with a predefined set of actions. Action Segmentation outputs can be fed into other systems for things like video-to-text and action localisation.

Given a video and a specific action to look for, action localisation attempts to pinpoint the precise frame at which the action is being carried out. What we call "\textbf{Action Localization}" is the process of pinpointing where and when something happens in a video. The x and y coordinates of an action can be extracted from a video using an action localisation model, which can determine in which frame an action begins and ends.

The goal of \textbf{temporal action detection (TAD)} is to locate each instance of action in an uncut video and assign it a semantic label and time interval. It's an essential part of video comprehension, and also one of the hardest parts.

In contrast to the widely studied problem of recognizing an action given a complete sequence, \textbf{action anticipation }aims to identify the action from only partially available videos.

In this section, we will discuss and classify different classification detection and Segmentation techniques for action recognition by transformers. Table 4 summarizes some recently emerges action recognition datasets and benchmarks and their details.

\section{The Performance Comparison, Bibliometric and Thematic Analysis}

A current trend is to compare the performance of deep learning methods on cloud-based leaderboards that update performance information related to challenging datasets in the field. They provide a fair comparison of performance in terms of standard metrics and keep on updating information about the best-performing models. Therefore, rather than setting up a new experimental setup to compare the action recognition performance of various transformer-based action recognition methods and their comparison with other CNN-based deep learning models, we share information about the best-performing models on standard benchmarking datasets.

 \begin{figure}[!tbp]
  \centering
  \begin{minipage}[b]{0.45\textwidth}
    \includegraphics[width=\textwidth]{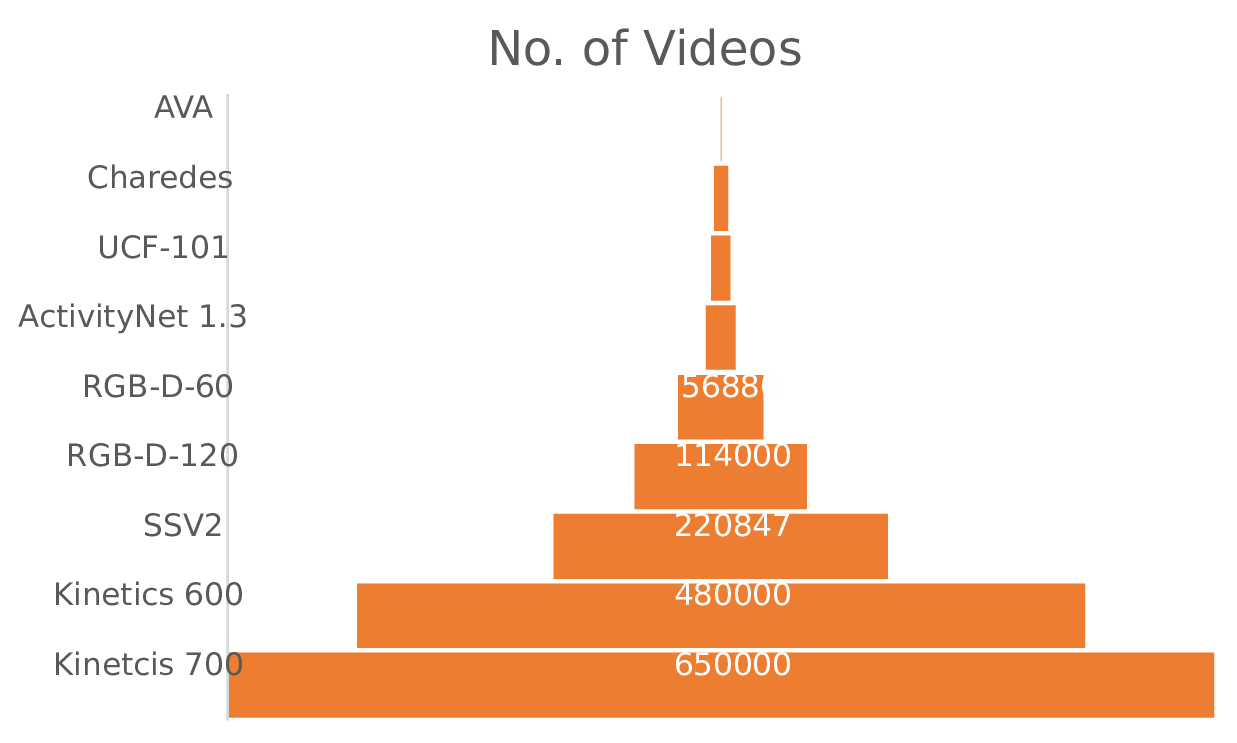}
     \end{minipage}
 \hfill
 \begin{minipage}[b]{0.45\textwidth}
    \includegraphics[width=\textwidth]{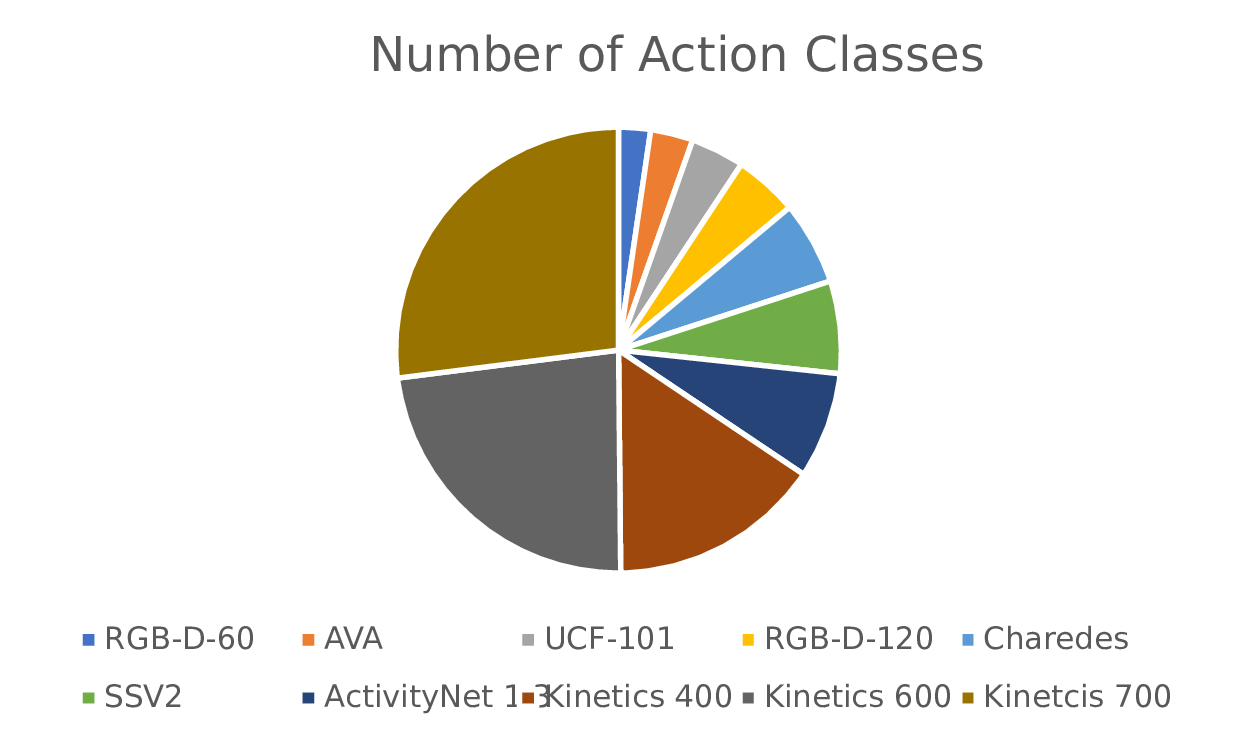}
 \end{minipage}
    \caption{The increasing size of action recognition datasets over time in terms of (left) number of action classes, and (b) the number of video clips. It shows a trend that the majority of the models are using trained or evaluated on very large video datasets. }
    \label{fig:DSize}
\end{figure}

\begin{figure}
\centering
\includegraphics[width=160mm]{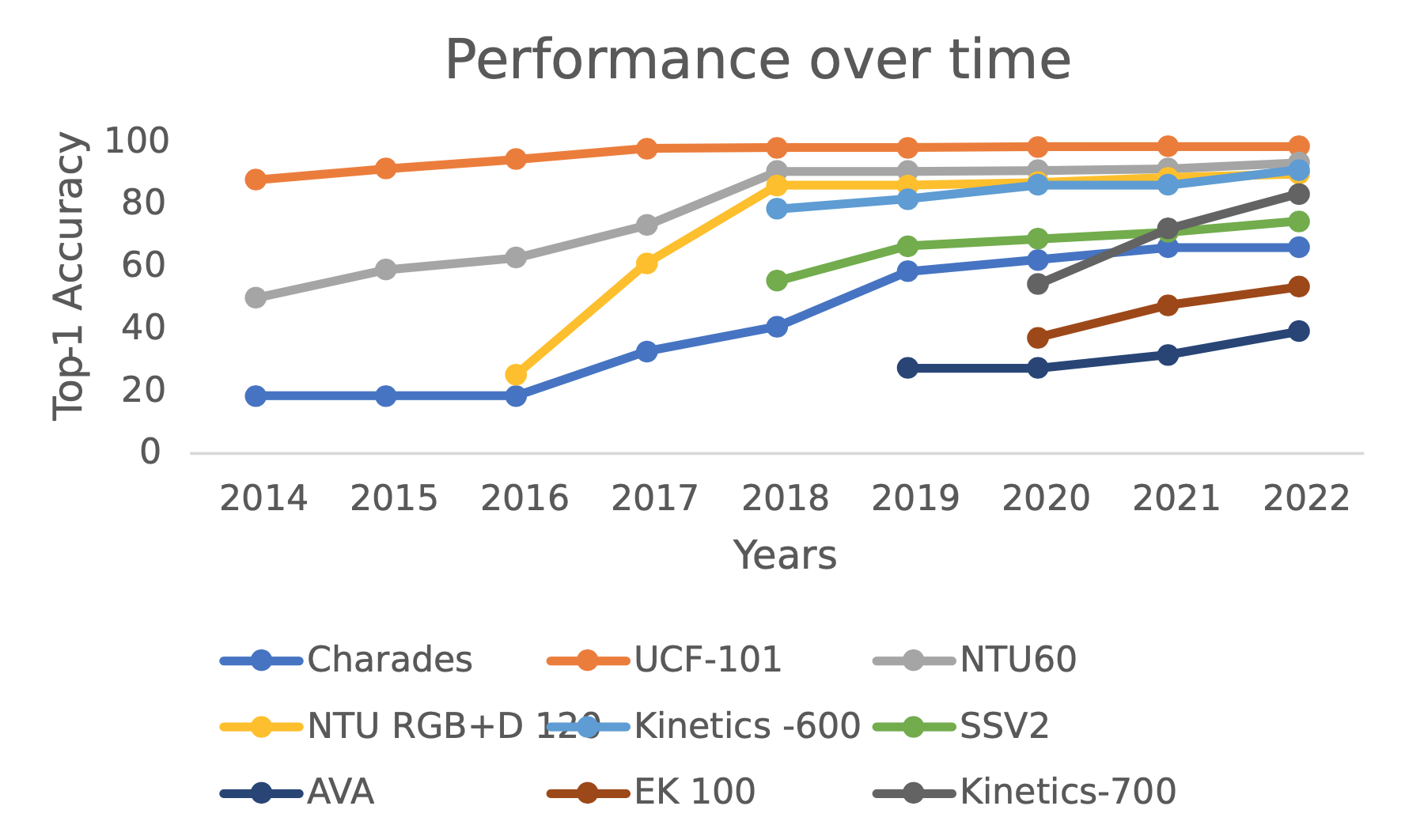} 
\caption{The improvement in the performance of top models ( in terms of Top-1 Accuracy) on different benchmark action detests. Traditional action datasets like UCF-101 have become saturated, so new datasets with large video sizes and specific action categories like Kitchen or indoor actions are merging. }
\label{fig:DACC}
\end{figure}

Top-1 accuracy is the conventional accuracy, and the model prediction (the one with the highest probability) must precisely match the anticipated outcome. It measures the proportion of instances where the predicted label corresponds to the single target label. Mean Average Precision (mAP) is a detection model evaluation metric. The mean of average precision (AP) values are computed using recall values between 0 and 1. Localization determines the location of an instance (e.g., bounding box coordinates), whereas classification identifies the instance's type. For each bounding box, the overlap between the predicted bounding box and the actual bounding box is determined. This is determined using IoU (intersection over union). AP is a weighted average of all categories. This is traditionally referred to as "mean average precision" (mAP).

The standard action recognition datasets that we have selected to compile this comparison include {Kinetics} \cite{Kinetics} and its version 400. 600 and 700 \cite{Kinetics,Kinetics600,Kinetics700}, Something-Something V2 \cite{Something}, Charades \cite{Charades}, ActivityNet \cite{Activitynet},NTU RGB+D \cite{Nturgb60}, NTU RGB+D 120 \cite{Nturgb120}, TVSeries \cite{TVseries}, 50 Salads \cite{50Salad}, Breakfast  \cite{Breakfast},  GTEA (Georgia Tech Egocentric Activity) \cite{GTEA} and EPIC-KITCHENS-100 \cite{EPIC-KITCHENS}.  Table 5 provides the list of top performing models on the leaderboard of challenging action recognition datasets: However, this table mentions only winner and runner-up approaches for each dataset.
 
One observation is their increasing size regarding the number of video data and the number of action classes. Fig.~\ref{fig:DSize} provides a visualization that gives insight into how action recognition datasets have increased in terms of action classes and number of video clips. Another observation is the recognition challenges posed by the data included in these datasets. Traditional action datasets like UCF-101 have become saturated, so new datasets with large class numbers and video sizes have emerged like Kinetics -700, which has 700 action categories, and EK-100, which has 50K action segments that provide a massive challenge for model evaluation.

\begin{table*}
\begin{center}
\begin{tabular}{|p{2cm}|l |l|l|l| l|l|l|l|}
\hline
\multicolumn{9}{ |c| }{{\cellcolor{green}} Representative Methodologies} \\
\hline
The Dataset used  & Rank & Model & Citation & Top-1 Acc & F1 & mAP &  GFLOPS & Transformer\\ \hline
\multirow{2}{*}{Kinetics 400} 
& 1  & MTV & \cite{MTV} & 89.9        &   -  &   -   & 73.57 T     &    \checkmark    \\
& 2  & CoCa & \cite{CoCa}                           & 88.9        &   -     &  -   &  -          &    \checkmark     \\
\hline
\multirow{2}{*}{Kinetics-600} 
& 1  & Merlot &  \cite{Merlot}& 91.1      &   -    &   -   & -      &    \checkmark    \\
& 2  & MTV & \cite{MTV}  & 90.3           &  -   &     -   &   &    \checkmark     \\
\hline
\multirow{2}{*}{SSV2} 
& 1  & VideoMAE & \cite{VideoMAE}& 75.4      &   -     &   - & $1436\times3$ &    \checkmark     \\
& 2  & MaskFeat & \cite{MaskFeat}  & 75.0      &   -      &  -   &    $2828\times3$       &    \checkmark     \\
\hline
\multirow{2}{*}{ActivityNet} 
& 1  & NS-Net &  \cite{SetN}&  90.2     &   -  &  94.3   & 26.1     &    \checkmark    \\
& 2  & TSQNet & \cite{TSQNet}  &  88.7    &   -     &   93.7   &    26.2      &    \checkmark     \\
\hline
\multirow{2}{*}{EK-100} 
& 1  & MM-Mix &  \cite{MM-Mix}&  53.6    &   -   &  -   & -     &    \checkmark    \\
& 2  & MTV & \cite{MTV}  &  50.5     &   -    &   -   &    -     &    \checkmark     \\
\hline
\multirow{2}{*}{NTU-60} 
& 1  & PoseC3D &  \cite{PoseC3D}& 94.1   &   -  &  -   &  0.6   & $\times $    \\
& 2  & 	HD-GCN & \cite{HD-GCN}  &  93.4    &   -    &   -   &  89.2 &    $\times$  \\
\hline
\multirow{2}{*}{NTU-120} 
& 1  & HD-GCN & \cite{HD-GCN}& 90.1     &   - &  -   & -     &    $\times $   \\
& 2  & 	HD-GCN & \cite{LST}  &  89.9     &   -   &   -   &    -     &    \checkmark    \\
\hline
\multirow{2}{*}{Charades} 
& 1  & CFNet & \cite{CFNet}& -  &   -   &  26.95  & -     &    $\times $   \\
& 2  & 	PDAN & \cite{PDAN}  &  -  &   -     &   26.5  &    -     &   $\times $  \\
\hline
\multirow{2}{*}{MIT} 
& 1  & Coca & \cite{CoCa}& 49.0    &   -  &  -  & -     &    $\times $    \\
& 2  & 	MTV & \cite{MTV}  &  47.2   &   -   &  -  &    81   &    $\times $   \\
\hline
\multirow{2}{*}{TVSeries} 
& 1  & GateHUB &\cite{GateHUB}& 89.6      &   -        &  -  & -     &    \checkmark    \\
& 2  & 	LST & \cite{LST}  &  89.1   &   -   &  -  &    -     &    \checkmark  \\
\hline
\multirow{2}{*}{Breakfast} 
& 1  & CETNet &  \cite{CETNet}& 74.9      &   61.9         &  -  & -     &    $\times $   \\
& 2  & 	U-Transformer & \cite{U-Transformer}  &  75   &   59.8    &  -  &    -     &    \checkmark   \\
\hline
\multirow{2}{*}{50 Salads} 
& 1  & Br-Prompt  & \cite{Br-Prompt}& 74.9      &   88.1         &  -  & -     &    \checkmark   \\
& 2  & 	CETNet & \cite{CETNet}  &  86.9     &   80.1     &  -  &    -     &   $\times $  \\
\hline
\multirow{2}{*}{GTEA} 
& 1  & Br-Prompt &  \cite{Br-Prompt}& 81.2     &   83.0          &  -  & -     &    \checkmark   \\
& 2  & 	DPRN & \cite{DPRN}  &  82.0    &   82.9     &  -  &    7.0  &    $\times $ \\
\hline
\end{tabular}
\captionsetup{singlelinecheck=false, font=small, labelfont=bf}
   \caption{ List of top performing models on the leaderboard of challenging action recognition datasets: However, this table mentions only winner and runner-up approaches for each dataset. To note, SSV2 is Something-something-V2, MIT is moments in time, EK-100 is Epic-Kitchen-100, NTU-60 is NTU-RGB-D-60 and NTU-120 is NTU-RGB-D-120 abbreviation. }

  \end{center}
    \end{table*}
By evaluating results on leaderboards, we discovered that vision transformer-based models dominate the action recognition performance in terms of accuracy. One example is Kinetics-600, for which all top 15 models are based on transformer and CNN, number 19 on the leaderboard. So that provides a good glimpse of the strength of transformers for action recognition. Fig.~\ref{fig:DACC} provides a graph to show the increase in performance on various  challenging action recognition datasets over time. It can be observed that top-1 accuracy on the traditional action datasets like UCF-101 have become saturated, so new datasets with large video sizes and specific action categories like Kitchen or indoor actions are merging.

Another comparison is in terms of efficiency. However, its accurate calculation is trying due to dependency on the underlying hardware. It is a common practice to use FLOPs to compare efficiency performance. However, in a recent evaluation of transformers, the authors argue that FLOPs are hardware-agnostic; it is not an accurate representation of real-world runtimes. On the other hand, latency gives a much more precise measure of efficiency as it runs the risk of changing over time due to hardware developments and CUDA  optimizations. 
 
\begin{figure}
\centering
\includegraphics[width=130mm]{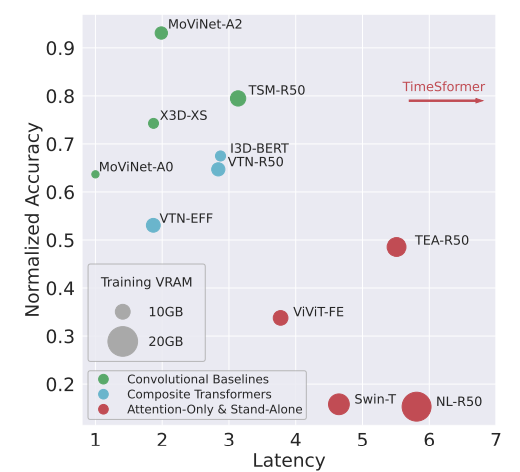} 
\caption{Pitt-falls of Action Transformers in terms of latency and accuracy compared to CNN models for lightweight action recognition. Image adapted from \cite{Evaluating}. }
\label{fig:DEff}
\end{figure}

  This particular paper provides a  comparison of various CNN and Transformer based approaches in terms of efficiency and accuracy comparison. However, this paper discussed that CNN models are still preferable and better in terms of efficacy. Fig.~\ref{fig:DEff} provides a  bubble chart that provides a comparison between the pure transformer and CNN approaches for action recognition in terms of latency and accuracy. Independent attention blocks also exhibit a sizeable latency overhead, which suggests that there is room for development in this area. It has been demonstrated that paying attention only is insufficient for the recognition of lightweight actions in general. Finding new ways to combine these architectural paradigms with other paradigms, such as what composite transformers do, is, therefore, a promising direction that should be pursued. The inconsistency and complexity of mobile latency results draw attention to the necessity of conducting in-depth research on mobile video transformers. Transformer models in vision frequently call for either an increase in the amount of data or an increase in the complexity of the training strategies. As a result, it is important to investigate whether or not redoing our latency and accuracy study with more extensive pre-training on a larger scale and more complex training strategies shifts the overall landscape of our findings.


 The inner dimensionality of self-attention is a promising target for improving transformer efficiency.  Stand-alone attention blocks show a considerable latency overhead, which could be worth improving. So, finding new ways to combine these architectures with other paradigms, as composite transformers do, is a promising direction. Therefore, while ViTs have achieved State-of-the-Art (SotA) performance on a variety of Computer Vision tasks, this does not guarantee that they will consistently outperform Convolutional Neural Networks (CNNs).

Unlike previous surveys that completely rely on manual analysis, we use automated visualizations of the bibliography for deeper literature analysis. It provides additional insights to understanding the original, chronological development, relevance, and impact of vision transformers on driving action recognition research. 

In the field of bibliometrics, visualization has proven to be an effective method for analyzing a wide range of networks, from those based on citations to those based on co-authorship or keyword co-occurrence. Density visualizations are useful for getting a high-level understanding of a bibliometric network's key components quickly. For textual data in English, it is possible to construct term co-occurrence networks using NLP methods. The motivation for using connected networks is to show the time and relevance-based evolution of the field. I will show how transformer work has emerged and influenced the design of vision transformers and video transformers ( please refer to Fig.~\ref{fig:Bib1}). In addition, to show evolutional connections, we analyze the existing literature related to vision transformer-based action recognition and try to analyze important directions of research. One way to analyze it ti to visualize the most saliant research work in this area. It can be done through bibliographic networks.

\begin{figure*}
\centering
\caption{Bibliometric Network for visualizing the evolution of vision transformers to Video Transformers. Similar papers have strong connecting lines and cluster together. Papers are arranged according to their similarity. Node size is the number of citations. Node color is the publishing year. Color dots inside three notable nodes are inserted to mention relevant models due to the large image size. The other closer and large nodes near the Vision transformer are the swin-transformer and other vision transformers. }
\label{fig:Bib1}
\end{figure*}

We have used VOSviewer \cite{VOSviewer}, which uses a distance-based approach to create a visual representation of any bibliometric network. Directed networks, such as those based on citation relationships, are categorized as undirected. VOSviewer automatically groups network nodes into clusters. A group of nodes that share many similar characteristics is called a cluster. A network's nodes are all neatly placed in one of the groups. VOSviewer uses color to represent a node's membership in a cluster in bibliometric network visualization. Waltman, Van Eck, and Noyons \cite{BIBNetworks} discuss the clustering method employed by VOSviewer. An algorithm for resolving an optimization problem is necessary for the method. VOSviewer employs the innovative MovingAlgo algorithm, developed by Waltman and Van Eck \cite{MovingAlgo}, for this function.

\begin{figure*}
\centering
\includegraphics[width=160 mm]{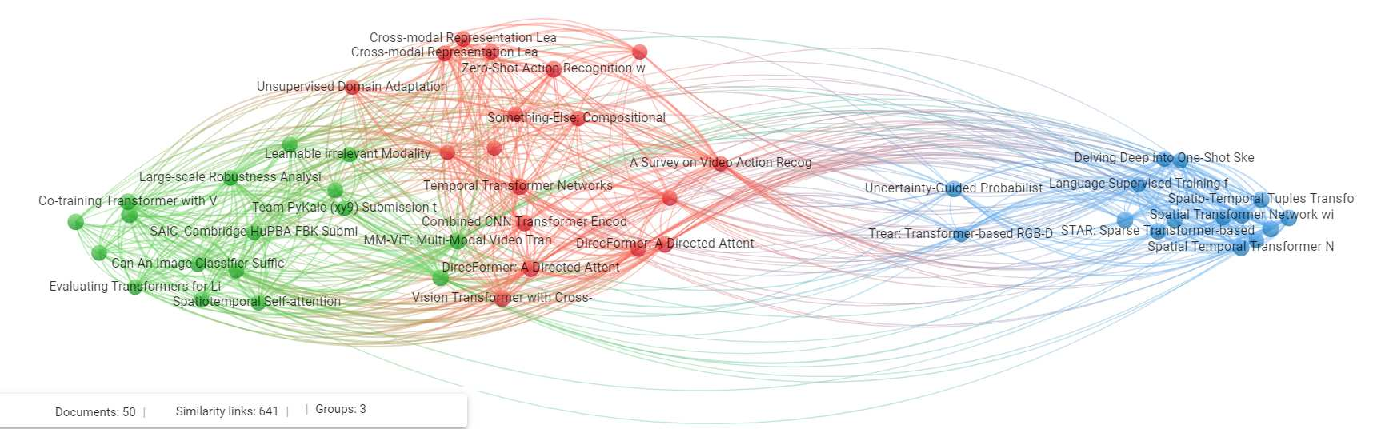} 
\caption{Visualization of Bibliometric Networks for action transformers" It shows the top 50 papers in terms of their relevance and impact on the topic and with each other. Each cluster provides a distinct theme and is represented by the same color dots. Mutual connections in the network are based on their term similarities. }
\label{fig:Bib2}
\end{figure*}

Another reason to use bibliographic networks was to find the most powerful theme related to transformer-based action recognition. For this purpose, we use a clustering approach. In cluster analysis, the number of sub-problems is set by the resolution. The greater the value of this parameter, the more clusters will be created. We tried to minimize the number of clusters to focus on the most representative work in terms of relevance ad impact, which resulted in only three clusters based on 50 research papers. Fig.~\ref{fig:Bib2} displays these clusters in three primary colors, and Table 1 lists five representative papers in each cluster. Such visualization of a bibliometric network provides an automated insight into relevant literature that can not be figured out manually. This visualization and its depth of understanding helped us to revise our taxonomies, which are discussed in the following sections. 

\begin{table}[H]
\setlength\extrarowheight{2mm}
    \centering
    \begin{tabularx}{\textwidth}{ | c | X |}
        \hline
         \textbf{{Cluster}} & \textbf{{Research Paper: The Title}} \\ \hline
 \statcirc{green} & Towards Training Stronger Video Vision Transformers for EPIC-KITCHENS-100 Action Recognition  \\ \hline    
 \statcirc{green} & Co-training Transformer with Videos and Images Improves Action Recognition  \\ \hline 
  \statcirc{green} & MM-ViT: Multi-Modal Video Transformer for Compressed Video Action Recognition  \\ \hline 
  \statcirc{green} & VideoLightFormer: Lightweight Action Recognition using Transformers   \\ \hline 
  \statcirc{green} & Can an Image Classifier Suffice For Action Recognition? \\ \hline 
  \statcirc{green} & Recurring the Transformer for Video Action Recognition  \\ \hline 
 \statcirc{blue} & IIP-Transformer: Intra-Inter-Part Transformer for Skeleton-Based Action Recognition   \\ \hline
 \statcirc{blue} & Hierarchical Graph Convolutional Skeleton Transformer for Action Recognition   \\ \hline
 \statcirc{blue} & Spatio-Temporal Tuples Transformer for Skeleton-Based Action Recognition   \\ \hline
 \statcirc{blue} & STAR: Sparse Transformer-based Action Recognition  \\ \hline
 \statcirc{blue} & Action Transformer: A Self-Attention Model for Short-Time Pose-Based Human Action Recognition   \\ \hline
 \statcirc{red} & DirecFormer: A Directed Attention in Transformer Approach to Robust Action Recognition \\ \hline
 \statcirc{red} & ActionCLIP: A New Paradigm for Video Action Recognition \\ \hline
 \statcirc{red} & Few-shot Action Recognition with Prototype-centered Attentive Learning \\ \hline
 \statcirc{red} & Cross-Modal Representation Learning for Zero-Shot Action Recognition \\ \hline
 \statcirc{red} & Zero-Shot Action Recognition with Transformer-based Video Semantic Embedding \\ \hline
 \end{tabularx}
\end{table}

\section{ Discussion and Future Directions}

The use of transformers is quickly becoming ubiquitous in various Computer Vision applications. Their greater modeling capacity, lower inductive biases, and global receptive fields allow them to outperform CNNs on large data sets in terms of accuracy. By decreasing receptive fields and adopting hierarchical pyramidal feature maps, modern, improved, and smaller ViTs like SWIN are essentially becoming CNN-integrated. When trained without knowledge distillation or additional data and when targeting lower accuracies, CNNs are still competitive with or superior to SotA ViTs on ImageNet. Some promising future research directions are as follows:

\begin{itemize}

\item Increasing the capacity of pure transformer models to model motion is a promising research course of action to take. It has been demonstrated by MViT \cite{MviT} that concentrating on the inner dimensionality of self-attention is a more promising target in terms of efficiency than progressive downsampling, encoder factorization, attention factorization, or linear self-attention.

\item An additional path would be to design a singular transformer that could accommodate multiple modalities. The 'OMNIVORE' model \cite{Omnivore} takes advantage of the adaptability offered by transformer-based architectures and is trained jointly on classification tasks drawn from a variety of modalities. It proposes a single model that is effective at classifying images, videos, and single-view 3D data by employing the same model parameters in all of these different classification tasks. Previous research has investigated the distinct visual modalities separately and developed distinct architectural frameworks for the recognition of images, videos, and three-dimensional data. OMNIVORE is easy to train, makes use of standard datasets that are readily available, and produces results that are on par with or even superior to those produced by modality-specific models of the same size. Another piece of work that is comparable is called Perceiver.

\item Transfer learning and Training strategy for video action transformers needs further exploration. Existing models use imagenet weights for pre-training. However, there has been less work put into figuring out how to train video transformers and their pre-training.  To begin, the performance of video transformers is improved by joint training on a variety of video datasets and label spaces (e.g., Kinetics is appearance-focused while SomethingSomething is motion-focused). Second, the video transformers acquire an even deeper understanding of video representations through additional co-training with images (in the form of single-frame videos). One such method is referred to as "Co-training Videos and Images for Action Recognition.

\item The most significant obstacle presented by the majority of Vision Transformer architectural designs is that they frequently call for an excessive number of tokens to produce satisfactory results. Even when using a 16x16 patch tokenization method, for example, a single 512x512 image is equivalent to 1024 tokens. When this occurs with videos that contain multiple frames, the output is tens of thousands of tokens because each video 'tubelet' (for example, a video segment that is 16 by 16 by 2) is converted into a token. In addition, because the outputs of one layer become the tokens for the next layer, the processing of a very large number of tokens is required at each layer. Considering that the computation required for a Transformer (and the amount of memory required) increases on a quadratic scale with the number of tokens, larger images, and longer videos may render Transformers unusable. This raises the question of whether or not it is truly necessary to process such a large number of tokens at each layer. A token Learner is a novel approach to visual representation learning that applies to both the task of understanding images and videos. It is based on a small number of tokens that have been learned adaptively. For the recognition tasks at hand, the objective is to acquire the skills necessary to learn how to isolate significant tokens from images and video frames. The acquisition of full Spatio-temporal tokens is one of the problems that still need to be solved. The current version of TokenLearner \cite{Tokenlearner}is geared towards locating spatial tokens across a series of frames; however, it is extensible and could be used to mine tokens across space-time volumes.
\end{itemize}

Few emerging areas are interesting including: 

1-\textbf{Action Quality Assessment}, also known as AQA, \cite{TPT} is essential for action comprehension, and resolving the task presents its own set of unique challenges as a result of subtle visual differences. In particular, performing an accurate analysis of the action quality has a tonne of potential in a wide variety of applications, including health care and the analysis of sports.

2- \textbf{Transformer based Zero-shot  Action Recognition} \cite{ResT}, is another popular direction direction. Its purpose is to categorize action classes that have not been encountered before in training. Traditionally, this is accomplished by training a network to map or regress visual inputs to a semantic space. The nearest neighbor classifier is then utilized to select the target class that is most closely related to the visual input.

3-\textbf{Compressed Video Action Recognition} \cite{MM-ViT} replaces optical flow with motion vectors present in the compressed video for encoding the movements of pixel blocks. Following up works achieve better performance by further exploiting residuals or refining motion vectors with the supervision of optical flow during training. Despite obtaining promising results, unlike our model, these methods ignore the rich inter-modal relations and fail to leverage audio signals.

Overall, the field is rapidly evolving, and its implementation and deployment on the edge and mobile devices will be interesting and challenging in the coming days.

\section{Conclusion}
	 In this survey, we provided analysis and summarisation of the key contributions and trends to adapting Transformers for visual recognition of human action.  Based on the existing literature, we tried to provide power taxonomies for action transformers based on their architecture, modality, and intended task, In particular, we discussed ways to embed and encode Spatio-temporal data, reduce its dimensionality, division into frame patches or Spatio-temporal cubes, and various representations for the recognition of human actions. We also provided an investigation of how Spatio-temporal attention has been optimized in the Transformer layer to handle longer sequences, typically by reducing the number of tokens in a single attention operation. Moreover, we investigated different network learning strategies such as self-supervised learning or zero-shot learning for transformer action recognition and associated losses. Finally, we provided a performance comparison of top approaches on different datasets and recommended design changes needed to improve their performance.

\bibliographystyle{unsrt}  

\begin{thebibliography}{1}

\bibitem{lipton2015critical}
Zachary~C Lipton, John Berkowitz, and Charles Elkan.
\newblock A critical review of recurrent neural networks for sequence learning.
\newblock {\em arXiv preprint arXiv:1506.00019}, 2015.

\bibitem{yang2020survey}
Shuoheng Yang, Yuxin Wang, and Xiaowen Chu.
\newblock A survey of deep learning techniques for neural machine translation.
\newblock {\em arXiv preprint arXiv:2002.07526}, 2020.

\bibitem{chen2020generative}
Mark Chen, Alec Radford, Rewon Child, Jeffrey Wu, Heewoo Jun, David Luan, and   Ilya Sutskever.
\newblock Generative pretraining from pixels.
\newblock In {\em International conference on machine learning}, pages
  1691--1703. PMLR, 2020.

\bibitem{zhang2019self}
Han Zhang, Ian Goodfellow, Dimitris Metaxas, and Augustus Odena.
\newblock Self-attention generative adversarial networks.
\newblock In {\em International conference on machine learning}, pages
  7354--7363. PMLR, 2019.

  

\bibitem{SurveyT1}
Salman Khan, Muzammal Naseer, Munawar Hayat, Syed~Waqas Zamir, Fahad~Shahbaz
  Khan, and Mubarak Shah.
\newblock Transformers in vision: A survey.
\newblock {\em ACM Computing Surveys (CSUR)}, 2021.

\bibitem{SurveyT2}
Tianyang Lin, Yuxin Wang, Xiangyang Liu, and Xipeng Qiu.
\newblock A survey of transformers.
\newblock {\em arXiv preprint arXiv:2106.04554}, 2021.

\bibitem{SurveyT3}
Kai Han, Yunhe Wang, Hanting Chen, Xinghao Chen, Jianyuan Guo, Zhenhua Liu,
  Yehui Tang, An~Xiao, Chunjing Xu, Yixing Xu, et~al.
\newblock A survey on vision transformer.
\newblock {\em IEEE transactions on pattern analysis and machine intelligence},
  2022.

\bibitem{RecentReview}
Preksha Pareek and Ankit Thakkar.
\newblock A survey on video-based human action recognition: recent updates,
  datasets, challenges, and applications.
\newblock {\em Artificial Intelligence Review}, 54(3):2259--2322, 2021.

\bibitem{ModalitySurvey}
Zehua Sun, Qiuhong Ke, Hossein Rahmani, Mohammed Bennamoun, Gang Wang, and Jun
  Liu.
\newblock Human action recognition from various data modalities: A review.
\newblock {\em IEEE Transactions on Pattern Analysis and Machine Intelligence},
  2022.

\bibitem{survey2022}
Yu~Kong and Yun Fu.
\newblock Human action recognition and prediction: A survey.
\newblock {\em International Journal of Computer Vision}, 130(5):1366--1401,
  2022.

\bibitem{Bahdanaul}
Dzmitry Bahdanau, Kyunghyun Cho, and Yoshua Bengio.
\newblock Neural machine translation by jointly learning to align and
  translate.
\newblock {\em arXiv preprint arXiv:1409.0473}, 2014.

\bibitem{Luong2015}
Minh-Thang Luong, Hieu Pham, and Christopher~D Manning.
\newblock Effective approaches to attention-based neural machine translation.
\newblock {\em arXiv preprint arXiv:1508.04025}, 2015.

\bibitem{XU}
Kelvin Xu, Jimmy Ba, Ryan Kiros, Kyunghyun Cho, Aaron Courville, Ruslan
  Salakhudinov, Rich Zemel, and Yoshua Bengio.
\newblock Show, attend and tell: Neural image caption generation with visual
  attention.
\newblock In {\em International conference on machine learning}, pages
  2048--2057. PMLR, 2015.

\bibitem{HAN}
Zichao Yang, Diyi Yang, Chris Dyer, Xiaodong He, Alex Smola, and Eduard Hovy.
\newblock Hierarchical attention networks for document classification.
\newblock In {\em Proceedings of the 2016 conference of the North American
  chapter of the association for computational linguistics: human language
  technologies}, pages 1480--1489, 2016.

\bibitem{Transformer}
Ashish Vaswani, Noam Shazeer, Niki Parmar, Jakob Uszkoreit, Llion Jones,
  Aidan~N Gomez, {\L}ukasz Kaiser, and Illia Polosukhin.
\newblock Attention is all you need.
\newblock {\em Advances in neural information processing systems}, 30, 2017.

\bibitem{GPT3}
Tom Brown, Benjamin Mann, Nick Ryder, Melanie Subbiah, Jared~D Kaplan, Prafulla
  Dhariwal, Arvind Neelakantan, Pranav Shyam, Girish Sastry, Amanda Askell,
  et~al.
\newblock Language models are few-shot learners.
\newblock {\em Advances in neural information processing systems},
  33:1877--1901, 2020.

\bibitem{Bert}
Jacob Devlin, Ming-Wei Chang, Kenton Lee, and Kristina Toutanova.
\newblock Bert: Pre-training of deep bidirectional transformers for language
  understanding.
\newblock {\em arXiv preprint arXiv:1810.04805}, 2018.

\bibitem{VIT}
Alexey Dosovitskiy, Lucas Beyer, Alexander Kolesnikov, Dirk Weissenborn,
  Xiaohua Zhai, Thomas Unterthiner, Mostafa Dehghani, Matthias Minderer, Georg
  Heigold, Sylvain Gelly, et~al.
\newblock An image is worth 16x16 words: Transformers for image recognition at
  scale.
\newblock {\em arXiv preprint arXiv:2010.11929}, 2020.

\bibitem{CNN}
Keiron O'Shea and Ryan Nash.
\newblock An introduction to convolutional neural networks.
\newblock {\em arXiv preprint arXiv:1511.08458}, 2015.

\bibitem{RNN}
Larry~R Medsker and LC~Jain.
\newblock Recurrent neural networks.
\newblock {\em Design and Applications}, 5:64--67, 2001.

\bibitem{Evaluating}
Raivo Koot, Markus Hennerbichler, and Haiping Lu.
\newblock Evaluating transformers for lightweight action recognition.
\newblock {\em arXiv preprint arXiv:2111.09641}, 2021.

\bibitem{Timeformer}
Gedas Bertasius, Heng Wang, and Lorenzo Torresani.
\newblock Is space-time attention all you need for video understanding?
\newblock In {\em ICML}, volume~2, page~4, 2021.

\bibitem{vivit}
Anurag Arnab, Mostafa Dehghani, Georg Heigold, Chen Sun, Mario Lu, and Cordelia
  Schmid.
\newblock Vivit: A video vision transformer.
\newblock In {\em Proceedings of the IEEE/CVF International Conference on
  Computer Vision}, pages 6836--6846, 2021.

\bibitem{VOSviewer}
Nees Van~Eck and Ludo Waltman.
\newblock Software survey: Vosviewer, a computer program for bibliometric
  mapping.
\newblock {\em scientometrics}, 84(2):523--538, 2010.

\bibitem{BIBNetworks}
Nees Jan~Van Eck and Ludo Waltman.
\newblock Visualizing bibliometric networks.
\newblock In {\em Measuring scholarly impact}, pages 285--320. Springer, 2014.

\bibitem{MovingAlgo}
Ludo Waltman and Nees~Jan Van~Eck.
\newblock A smart local moving algorithm for large-scale modularity-based
  community detection.
\newblock {\em The European physical journal B}, 86(11):1--14, 2013.

\bibitem{LSTM}
Ralf~C Staudemeyer and Eric~Rothstein Morris.
\newblock Understanding lstm--a tutorial into long short-term memory recurrent
  neural networks.
\newblock {\em arXiv preprint arXiv:1909.09586}, 2019.

\bibitem{COVER}
Bowen Zhang, Jiahui Yu, Christopher Fifty, Wei Han, Andrew~M Dai, Ruoming Pang,
  and Fei Sha.
\newblock Co-training transformer with videos and images improves action
  recognition.
\newblock {\em arXiv preprint arXiv:2112.07175}, 2021.

\bibitem{Actionformer}
Chenlin Zhang, Jianxin Wu, and Yin Li.
\newblock Actionformer: Localizing moments of actions with transformers.
\newblock {\em arXiv preprint arXiv:2202.07925}, 2022.

\bibitem{AVA}
Chunhui Gu, Chen Sun, David~A Ross, Carl Vondrick, Caroline Pantofaru, Yeqing
  Li, Sudheendra Vijayanarasimhan, George Toderici, Susanna Ricco, Rahul
  Sukthankar, et~al.
\newblock Ava: A video dataset of spatio-temporally localized atomic visual
  actions.
\newblock In {\em Proceedings of the IEEE Conference on Computer Vision and
  Pattern Recognition}, pages 6047--6056, 2018.

\bibitem{STAN}
Edward Fish, Jon Weinbren, and Andrew Gilbert.
\newblock Two-stream transformer architecture for long video understanding.
\newblock {\em arXiv preprint arXiv:2208.01753}, 2022.

\bibitem{VTCE}
Mei~Chee Leong, Haosong Zhang, Hui~Li Tan, Liyuan Li, and Joo~Hwee Lim.
\newblock Combined cnn transformer encoder for enhanced fine-grained human
  action recognition.
\newblock {\em arXiv preprint arXiv:2208.01897}, 2022.

\bibitem{FRAB}
M~Kalfaoglu, Sinan Kalkan, and A~Aydin Alatan.
\newblock Late temporal modeling in 3d cnn architectures with bert for action
  recognition.
\newblock In {\em European Conference on Computer Vision}, pages 731--747.
  Springer, 2020.

\bibitem{TadTR}
Xiaolong Liu, Qimeng Wang, Yao Hu, Xu~Tang, Shiwei Zhang, Song Bai, and Xiang
  Bai.
\newblock End-to-end temporal action detection with transformer.
\newblock {\em IEEE Transactions on Image Processing}, 2022.

\bibitem{CBT}
Chen Sun, Fabien Baradel, Kevin Murphy, and Cordelia Schmid.
\newblock Learning video representations using contrastive bidirectional
  transformer.
\newblock {\em arXiv preprint arXiv:1906.05743}, 2019.

\bibitem{MS-TCT}
Rui Dai, Srijan Das, Kumara Kahatapitiya, Michael~S Ryoo, and Francois Bremond.
\newblock Ms-tct: Multi-scale temporal convtransformer for action detection.
\newblock In {\em Proceedings of the IEEE/CVF Conference on Computer Vision and
  Pattern Recognition}, pages 20041--20051, 2022.

\bibitem{ST-TR}
Chiara Plizzari, Marco Cannici, and Matteo Matteucci.
\newblock Spatial temporal transformer network for skeleton-based action
  recognition.
\newblock In {\em International Conference on Pattern Recognition}, pages
  694--701. Springer, 2021.

\bibitem{VTN}
Alexander Kozlov, Vadim Andronov, and Yana Gritsenko.
\newblock Lightweight network architecture for real-time action recognition.
\newblock In {\em Proceedings of the 35th Annual ACM Symposium on Applied
  Computing}, pages 2074--2080, 2020.

\bibitem{TPT}
Yang Bai, Desen Zhou, Songyang Zhang, Jian Wang, Errui Ding, Yu~Guan, Yang
  Long, and Jingdong Wang.
\newblock Action quality assessment with temporal parsing transformer.
\newblock {\em arXiv preprint arXiv:2207.09270}, 2022.

\bibitem{VideoLightFormer}
Raivo Koot and Haiping Lu.
\newblock Videolightformer: Lightweight action recognition using transformers.
\newblock {\em arXiv preprint arXiv:2107.00451}, 2021.

\bibitem{Actor-Transformer}
Kirill Gavrilyuk, Ryan Sanford, Mehrsan Javan, and Cees~GM Snoek.
\newblock Actor-transformers for group activity recognition.
\newblock In {\em Proceedings of the IEEE/CVF Conference on Computer Vision and
  Pattern Recognition}, pages 839--848, 2020.

\bibitem{UGPT}
Hongji Guo, Hanjing Wang, and Qiang Ji.
\newblock Uncertainty-guided probabilistic transformer for complex action
  recognition.
\newblock In {\em Proceedings of the IEEE/CVF Conference on Computer Vision and
  Pattern Recognition}, pages 20052--20061, 2022.

\bibitem{VATN}
Rohit Girdhar, Joao Carreira, Carl Doersch, and Andrew Zisserman.
\newblock Video action transformer network.
\newblock In {\em Proceedings of the IEEE/CVF conference on computer vision and
  pattern recognition}, pages 244--253, 2019.

\bibitem{KA-AGTN}
Yanan Liu, Hao Zhang, Dan Xu, and Kangjian He.
\newblock Graph transformer network with temporal kernel attention for
  skeleton-based action recognition.
\newblock {\em Knowledge-Based Systems}, 240:108146, 2022.

\bibitem{RTDNet}
Jing Tan, Jiaqi Tang, Limin Wang, and Gangshan Wu.
\newblock Relaxed transformer decoders for direct action proposal generation.
\newblock In {\em Proceedings of the IEEE/CVF International Conference on
  Computer Vision}, pages 13526--13535, 2021.

\bibitem{STAT}
Bonan Li, Pengfei Xiong, Congying Han, and Tiande Guo.
\newblock Shrinking temporal attention in transformers for video action
  recognition.
\newblock 2022.

\bibitem{Vidtr}
Yanyi Zhang, Xinyu Li, Chunhui Liu, Bing Shuai, Yi~Zhu, Biagio Brattoli, Hao
  Chen, Ivan Marsic, and Joseph Tighe.
\newblock Vidtr: Video transformer without convolutions.
\newblock In {\em Proceedings of the IEEE/CVF International Conference on
  Computer Vision}, pages 13577--13587, 2021.

\bibitem{MTV}
Shen Yan, Xuehan Xiong, Anurag Arnab, Zhichao Lu, Mi~Zhang, Chen Sun, and
  Cordelia Schmid.
\newblock Multiview transformers for video recognition.
\newblock In {\em Proceedings of the IEEE/CVF Conference on Computer Vision and
  Pattern Recognition}, pages 3333--3343, 2022.

\bibitem{STPT}
Yuetian Weng, Zizheng Pan, Mingfei Han, Xiaojun Chang, and Bohan Zhuang.
\newblock An efficient spatio-temporal pyramid transformer for action
  detection.
\newblock {\em arXiv preprint arXiv:2207.10448}, 2022.

\bibitem{ASFormer}
Fangqiu Yi, Hongyu Wen, and Tingting Jiang.
\newblock Asformer: Transformer for action segmentation.
\newblock {\em arXiv preprint arXiv:2110.08568}, 2021.

\bibitem{AcT}
Vittorio Mazzia, Simone Angarano, Francesco Salvetti, Federico Angelini, and
  Marcello Chiaberge.
\newblock Action transformer: A self-attention model for short-time pose-based
  human action recognition.
\newblock {\em Pattern Recognition}, 124:108487, 2022.

\bibitem{DVT}
Jue Wang and Lorenzo Torresani.
\newblock Deformable video transformer.
\newblock In {\em Proceedings of the IEEE/CVF Conference on Computer Vision and
  Pattern Recognition}, pages 14053--14062, 2022.

\bibitem{V-SWIN}
Ze~Liu, Jia Ning, Yue Cao, Yixuan Wei, Zheng Zhang, Stephen Lin, and Han Hu.
\newblock Video swin transformer.
\newblock In {\em Proceedings of the IEEE/CVF Conference on Computer Vision and
  Pattern Recognition}, pages 3202--3211, 2022.

\bibitem{RViT}
Jiewen Yang, Xingbo Dong, Liujun Liu, Chao Zhang, Jiajun Shen, and Dahai Yu.
\newblock Recurring the transformer for video action recognition.
\newblock In {\em Proceedings of the IEEE/CVF Conference on Computer Vision and
  Pattern Recognition}, pages 14063--14073, 2022.

\bibitem{DirecFormer}
Thanh-Dat Truong, Quoc-Huy Bui, Chi~Nhan Duong, Han-Seok Seo, Son~Lam Phung,
  Xin Li, and Khoa Luu.
\newblock Direcformer: A directed attention in transformer approach to robust
  action recognition.
\newblock In {\em Proceedings of the IEEE/CVF Conference on Computer Vision and
  Pattern Recognition}, pages 20030--20040, 2022.

\bibitem{STAR}
Feng Shi, Chonghan Lee, Liang Qiu, Yizhou Zhao, Tianyi Shen, Shivran
  Muralidhar, Tian Han, Song-Chun Zhu, and Vijaykrishnan Narayanan.
\newblock Star: Sparse transformer-based action recognition.
\newblock {\em arXiv preprint arXiv:2107.07089}, 2021.

\bibitem{AGT}
Megha Nawhal and Greg Mori.
\newblock Activity graph transformer for temporal action localization.
\newblock {\em arXiv preprint arXiv:2101.08540}, 2021.

\bibitem{SGT}
Yanan Liu, Hao Zhang, Dan Xu, and Kangjian He.
\newblock Graph transformer network with temporal kernel attention for
  skeleton-based action recognition.
\newblock {\em Knowledge-Based Systems}, 240:108146, 2022.

\bibitem{MM-ViT}
Jiawei Chen and Chiu~Man Ho.
\newblock Mm-vit: Multi-modal video transformer for compressed video action
  recognition.
\newblock In {\em Proceedings of the IEEE/CVF Winter Conference on Applications
  of Computer Vision}, pages 1910--1921, 2022.

\bibitem{Realistic}
Khurram Soomro and Amir~R Zamir.
\newblock Action recognition in realistic sports videos.
\newblock In {\em Computer vision in sports}, pages 181--208. Springer, 2014.

\bibitem{Drivering}
Mingqi Lu, Yaocong Hu, and Xiaobo Lu.
\newblock Driver action recognition using deformable and dilated faster r-cnn
  with optimized region proposals.
\newblock {\em Applied Intelligence}, 50(4):1100--1111, 2020.

\bibitem{surveillance}
Weiyao Lin, Ming-Ting Sun, Radha Poovandran, and Zhengyou Zhang.
\newblock Human activity recognition for video surveillance.
\newblock In {\em 2008 IEEE International Symposium on Circuits and Systems
  (ISCAS)}, pages 2737--2740. IEEE, 2008.

\bibitem{Pose2}
Jamie Shotton, Andrew Fitzgibbon, Mat Cook, Toby Sharp, Mark Finocchio, Richard
  Moore, Alex Kipman, and Andrew Blake.
\newblock Real-time human pose recognition in parts from single depth images.
\newblock In {\em CVPR 2011}, pages 1297--1304. Ieee, 2011.

\bibitem{posep}
Ke~Sun, Bin Xiao, Dong Liu, and Jingdong Wang.
\newblock Deep high-resolution representation learning for human pose
  estimation.
\newblock In {\em Proceedings of the IEEE/CVF conference on computer vision and
  pattern recognition}, pages 5693--5703, 2019.

\bibitem{CoCa}
Jiahui Yu, Zirui Wang, Vijay Vasudevan, Legg Yeung, Mojtaba Seyedhosseini, and
  Yonghui Wu.
\newblock Coca: Contrastive captioners are image-text foundation models.
\newblock {\em arXiv preprint arXiv:2205.01917}, 2022.

\bibitem{Merlot}
Rowan Zellers, Jiasen Lu, Ximing Lu, Youngjae Yu, Yanpeng Zhao, Mohammadreza
  Salehi, Aditya Kusupati, Jack Hessel, Ali Farhadi, and Yejin Choi.
\newblock Merlot reserve: Neural script knowledge through vision and language
  and sound.
\newblock In {\em Proceedings of the IEEE/CVF Conference on Computer Vision and
  Pattern Recognition}, pages 16375--16387, 2022.

\bibitem{AAA}
Yunhua Zhang, Hazel Doughty, Ling Shao, and Cees~GM Snoek.
\newblock Audio-adaptive activity recognition across video domains.
\newblock In {\em Proceedings of the IEEE/CVF Conference on Computer Vision and
  Pattern Recognition}, pages 13791--13800, 2022.

\bibitem{EAO}
Nina Shvetsova, Brian Chen, Andrew Rouditchenko, Samuel Thomas, Brian
  Kingsbury, Rogerio~S Feris, David Harwath, James Glass, and Hilde Kuehne.
\newblock Everything at once-multi-modal fusion transformer for video
  retrieval.
\newblock In {\em Proceedings of the IEEE/CVF Conference on Computer Vision and
  Pattern Recognition}, pages 20020--20029, 2022.

\bibitem{CCD}
Zhengyuan Yang, Jingen Liu, Jing Huang, Xiaodong He, Tao Mei, Chenliang Xu, and
  Jiebo Luo.
\newblock Cross-modal contrastive distillation for instructional activity
  anticipation.
\newblock {\em arXiv preprint arXiv:2201.06734}, 2022.

\bibitem{Actionclip}
Mengmeng Wang, Jiazheng Xing, and Yong Liu.
\newblock Actionclip: A new paradigm for video action recognition.
\newblock {\em arXiv preprint arXiv:2109.08472}, 2021.

\bibitem{LST}
Wangmeng Xiang, Chao Li, Yuxuan Zhou, Biao Wang, and Lei Zhang.
\newblock Language supervised training for skeleton-based action recognition.
\newblock {\em arXiv preprint arXiv:2208.05318}, 2022.

\bibitem{speech2action}
Arsha Nagrani, Chen Sun, David Ross, Rahul Sukthankar, Cordelia Schmid, and
  Andrew Zisserman.
\newblock Speech2action: Cross-modal supervision for action recognition.
\newblock In {\em Proceedings of the IEEE/CVF conference on computer vision and
  pattern recognition}, pages 10317--10326, 2020.

\bibitem{MM-Mix}
Xuehan Xiong, Anurag Arnab, Arsha Nagrani, and Cordelia Schmid.
\newblock M\&m mix: A multimodal multiview transformer ensemble.
\newblock {\em arXiv preprint arXiv:2206.09852}, 2022.

\bibitem{LOSOCV}
Momal Ijaz, Renato Diaz, and Chen Chen.
\newblock Multimodal transformer for nursing activity recognition.
\newblock In {\em Proceedings of the IEEE/CVF Conference on Computer Vision and
  Pattern Recognition}, pages 2065--2074, 2022.

\bibitem{Vatt}
Hassan Akbari, Liangzhe Yuan, Rui Qian, Wei-Hong Chuang, Shih-Fu Chang, Yin
  Cui, and Boqing Gong.
\newblock Vatt: Transformers for multimodal self-supervised learning from raw
  video, audio and text.
\newblock {\em Advances in Neural Information Processing Systems},
  34:24206--24221, 2021.

\bibitem{EAMAT}
Shuo Yang and Xinxiao Wu.
\newblock Entity-aware and motion-aware transformers for language-driven action
  localization in videos.
\newblock {\em arXiv preprint arXiv:2205.05854}, 2022.

\bibitem{MTCN}
Evangelos Kazakos, Jaesung Huh, Arsha Nagrani, Andrew Zisserman, and Dima
  Damen.
\newblock With a little help from my temporal context: Multimodal egocentric
  action recognition.
\newblock {\em arXiv preprint arXiv:2111.01024}, 2021.

\bibitem{UDAVT}
Victor G~Turrisi da~Costa, Giacomo Zara, Paolo Rota, Thiago Oliveira-Santos,
  Nicu Sebe, Vittorio Murino, and Elisa Ricci.
\newblock Unsupervised domain adaptation for video transformers in action
  recognition.
\newblock {\em arXiv preprint arXiv:2207.12842}, 2022.

\bibitem{Lin.Attn.Transf.}
John Ridley, Huseyin Coskun, David~Joseph Tan, Nassir Navab, and Federico
  Tombari.
\newblock Transformers in action: Weakly supervised action segmentation.
\newblock {\em arXiv preprint arXiv:2201.05675}, 2022.

\bibitem{U-Transformer}
Dazhao Du, Bing Su, Yu~Li, Zhongang Qi, Lingyu Si, and Ying Shan.
\newblock Efficient u-transformer with boundary-aware loss for action
  segmentation.
\newblock {\em arXiv preprint arXiv:2205.13425}, 2022.

\bibitem{CETNet}
Jiahui Wang, Zhenyou Wang, Shanna Zhuang, and Hui Wang.
\newblock Cross-enhancement transformer for action segmentation.
\newblock {\em arXiv preprint arXiv:2205.09445}, 2022.

\bibitem{nRM}
Simon H{\"a}ring, Raphael Memmesheimer, and Dietrich Paulus.
\newblock Action segmentation on representations of skeleton sequences using
  transformer networks.
\newblock In {\em 2021 IEEE International Conference on Image Processing
  (ICIP)}, pages 3053--3057. IEEE, 2021.

\bibitem{SCT}
Mohsen Fayyaz and Jurgen Gall.
\newblock Sct: Set constrained temporal transformer for set supervised action
  segmentation.
\newblock In {\em Proceedings of the IEEE/CVF Conference on Computer Vision and
  Pattern Recognition}, pages 501--510, 2020.

\bibitem{TALFramework}
Zhiwu Qing, Xiang Wang, Ziyuan Huang, Yutong Feng, Shiwei Zhang, Mingqian Tang,
  Changxin Gao, Nong Sang, et~al.
\newblock Exploring stronger feature for temporal action localization.
\newblock {\em arXiv preprint arXiv:2106.13014}, 2021.

\bibitem{Ag-Trans}
Peisen Zhao, Lingxi Xie, Ya~Zhang, and Qi~Tian.
\newblock Actionness-guided transformer for anchor-free temporal action
  localization.
\newblock {\em IEEE Signal Processing Letters}, 29:194--198, 2021.

\bibitem{TubeR}
Jiaojiao Zhao, Yanyi Zhang, Xinyu Li, Hao Chen, Bing Shuai, Mingze Xu, Chunhui
  Liu, Kaustav Kundu, Yuanjun Xiong, Davide Modolo, et~al.
\newblock Tuber: Tubelet transformer for video action detection.
\newblock In {\em Proceedings of the IEEE/CVF Conference on Computer Vision and
  Pattern Recognition}, pages 13598--13607, 2022.

\bibitem{Oadtr}
Xiang Wang, Shiwei Zhang, Zhiwu Qing, Yuanjie Shao, Zhengrong Zuo, Changxin
  Gao, and Nong Sang.
\newblock Oadtr: Online action detection with transformers.
\newblock In {\em Proceedings of the IEEE/CVF International Conference on
  Computer Vision}, pages 7565--7575, 2021.

\bibitem{Stargazer}
Junwei Liang, He~Zhu, Enwei Zhang, and Jun Zhang.
\newblock Stargazer: A transformer-based driver action detection system for
  intelligent transportation.
\newblock In {\em Proceedings of the IEEE/CVF Conference on Computer Vision and
  Pattern Recognition}, pages 3160--3167, 2022.

\bibitem{HORST}
Tsung-Ming Tai, Giuseppe Fiameni, Cheng-Kuang Lee, and Oswald Lanz.
\newblock Higher order recurrent space-time transformer for video action
  prediction.
\newblock {\em arXiv preprint arXiv:2104.08665}, 2021.

\bibitem{URM-CTP}
Tsung-Ming Tai, Giuseppe Fiameni, Cheng-Kuang Lee, Simon See, and Oswald Lanz.
\newblock Unified recurrence modeling for video action anticipation.
\newblock {\em arXiv preprint arXiv:2206.01009}, 2022.

\bibitem{MM-Transformer}
Debaditya Roy and Basura Fernando.
\newblock Action anticipation using pairwise human-object interactions and
  transformers.
\newblock {\em IEEE Transactions on Image Processing}, 30:8116--8129, 2021.

\bibitem{AVT}
Rohit Girdhar and Kristen Grauman.
\newblock Anticipative video transformer.
\newblock In {\em Proceedings of the IEEE/CVF International Conference on
  Computer Vision}, pages 13505--13515, 2021.

\bibitem{FUTR}
Dayoung Gong, Joonseok Lee, Manjin Kim, Seong~Jong Ha, and Minsu Cho.
\newblock Future transformer for long-term action anticipation.
\newblock In {\em Proceedings of the IEEE/CVF Conference on Computer Vision and
  Pattern Recognition}, pages 3052--3061, 2022.

\bibitem{Kinetics}
Will Kay, Joao Carreira, Karen Simonyan, Brian Zhang, Chloe Hillier, Sudheendra
  Vijayanarasimhan, Fabio Viola, Tim Green, Trevor Back, Paul Natsev, et~al.
\newblock The kinetics human action video dataset.
\newblock {\em arXiv preprint arXiv:1705.06950}, 2017.

\bibitem{Kinetics600}
Joao Carreira, Eric Noland, Andras Banki-Horvath, Chloe Hillier, and Andrew
  Zisserman.
\newblock A short note about c.
\newblock {\em arXiv preprint arXiv:1808.01340}, 2018.

\bibitem{Kinetics700}
Joao Carreira, Eric Noland, Chloe Hillier, and Andrew Zisserman.
\newblock A short note on the kinetics-700 human action dataset.
\newblock {\em arXiv preprint arXiv:1907.06987}, 2019.

\bibitem{Something}
Raghav Goyal, Samira Ebrahimi~Kahou, Vincent Michalski, Joanna Materzynska,
  Susanne Westphal, Heuna Kim, Valentin Haenel, Ingo Fruend, Peter Yianilos,
  Moritz Mueller-Freitag, et~al.
\newblock The" something something" video database for learning and evaluating
  visual common sense.
\newblock In {\em Proceedings of the IEEE international conference on computer
  vision}, pages 5842--5850, 2017.

\bibitem{Charades}
Gunnar~A Sigurdsson, G{\"u}l Varol, Xiaolong Wang, Ali Farhadi, Ivan Laptev,
  and Abhinav Gupta.
\newblock Hollywood in homes: Crowdsourcing data collection for activity
  understanding.
\newblock In {\em European Conference on Computer Vision}, pages 510--526.
  Springer, 2016.

\bibitem{VideoMAE}
Zhan Tong, Yibing Song, Jue Wang, and Limin Wang.
\newblock Videomae: Masked autoencoders are data-efficient learners for
  self-supervised video pre-training.
\newblock {\em arXiv preprint arXiv:2203.12602}, 2022.

\bibitem{MaskFeat}
Chen Wei, Haoqi Fan, Saining Xie, Chao-Yuan Wu, Alan Yuille, and Christoph
  Feichtenhofer.
\newblock Masked feature prediction for self-supervised visual pre-training.
\newblock In {\em Proceedings of the IEEE/CVF Conference on Computer Vision and
  Pattern Recognition}, pages 14668--14678, 2022.

\bibitem{SetN}
Boyang Xia, Wenhao Wu, Haoran Wang, Rui Su, Dongliang He, Haosen Yang, Xiaoran
  Fan, and Wanli Ouyang.
\newblock Nsnet: Non-saliency suppression sampler for efficient video
  recognition.
\newblock {\em arXiv preprint arXiv:2207.10388}, 2022.

\bibitem{TSQNet}
Boyang Xia, Zhihao Wang, Wenhao Wu, Haoran Wang, and Jungong Han.
\newblock Temporal saliency query network for efficient video recognition.
\newblock {\em arXiv preprint arXiv:2207.10379}, 2022.

\bibitem{PoseC3D}
Haodong Duan, Yue Zhao, Kai Chen, Dahua Lin, and Bo~Dai.
\newblock Revisiting skeleton-based action recognition.
\newblock In {\em Proceedings of the IEEE/CVF Conference on Computer Vision and
  Pattern Recognition}, pages 2969--2978, 2022.

\bibitem{HD-GCN}
Jialin Gao, Tong He, Xi~Zhou, and Shiming Ge.
\newblock Skeleton-based action recognition with focusing-diffusion graph
  convolutional networks.
\newblock {\em IEEE Signal Processing Letters}, 28:2058--2062, 2021.

\bibitem{CFNet}
Kumara Kahatapitiya, Zhou Ren, Haoxiang Li, Zhenyu Wu, and Michael~S Ryoo.
\newblock Self-supervised pretraining with classification labels for temporal
  activity detection.
\newblock {\em arXiv preprint arXiv:2111.13675}, 2021.

\bibitem{PDAN}
Rui Dai, Srijan Das, Luca Minciullo, Lorenzo Garattoni, Gianpiero Francesca,
  and Francois Bremond.
\newblock Pdan: Pyramid dilated attention network for action detection.
\newblock In {\em Proceedings of the IEEE/CVF Winter Conference on Applications
  of Computer Vision}, pages 2970--2979, 2021.

\bibitem{GateHUB}
Junwen Chen, Gaurav Mittal, Ye~Yu, Yu~Kong, and Mei Chen.
\newblock Gatehub: Gated history unit with background suppression for online
  action detection.
\newblock In {\em Proceedings of the IEEE/CVF Conference on Computer Vision and
  Pattern Recognition}, pages 19925--19934, 2022.

\bibitem{Br-Prompt}
Muheng Li, Lei Chen, Yueqi Duan, Zhilan Hu, Jianjiang Feng, Jie Zhou, and Jiwen
  Lu.
\newblock Bridge-prompt: Towards ordinal action understanding in instructional
  videos.
\newblock In {\em Proceedings of the IEEE/CVF Conference on Computer Vision and
  Pattern Recognition}, pages 19880--19889, 2022.

\bibitem{DPRN}
Junyong Park, Daekyum Kim, Sejoon Huh, and Sungho Jo.
\newblock Maximization and restoration: Action segmentation through dilation
  passing and temporal reconstruction.
\newblock {\em Pattern Recognition}, 129:108764, 2022.

\bibitem{Activitynet}
Fabian Caba~Heilbron, Victor Escorcia, Bernard Ghanem, and Juan Carlos~Niebles.
\newblock Activitynet: A large-scale video benchmark for human activity
  understanding.
\newblock In {\em Proceedings of the ieee conference on computer vision and
  pattern recognition}, pages 961--970, 2015.

\bibitem{Moments}
Mathew Monfort, Alex Andonian, Bolei Zhou, Kandan Ramakrishnan, Sarah~Adel
  Bargal, Tom Yan, Lisa Brown, Quanfu Fan, Dan Gutfreund, Carl Vondrick, et~al.
\newblock Moments in time dataset: one million videos for event understanding.
\newblock {\em IEEE transactions on pattern analysis and machine intelligence},
  42(2):502--508, 2019.

\bibitem{Nturgb60}
Amir Shahroudy, Jun Liu, Tian-Tsong Ng, and Gang Wang.
\newblock Ntu rgb+ d: A large scale dataset for 3d human activity analysis.
\newblock In {\em Proceedings of the IEEE conference on computer vision and
  pattern recognition}, pages 1010--1019, 2016.

\bibitem{Nturgb120}
Jun Liu, Amir Shahroudy, Mauricio Perez, Gang Wang, Ling-Yu Duan, and Alex~C
  Kot.
\newblock Ntu rgb+ d 120: A large-scale benchmark for 3d human activity
  understanding.
\newblock {\em IEEE transactions on pattern analysis and machine intelligence},
  42(10):2684--2701, 2019.

\bibitem{TVseries}
Roeland~De Geest, Efstratios Gavves, Amir Ghodrati, Zhenyang Li, Cees Snoek,
  and Tinne Tuytelaars.
\newblock Online action detection.
\newblock In {\em European Conference on Computer Vision}, pages 269--284.
  Springer, 2016.

\bibitem{50Salad}
Sebastian Stein and Stephen~J McKenna.
\newblock Recognising complex activities with histograms of relative tracklets.
\newblock {\em Computer Vision and Image Understanding}, 154:82--93, 2017.

\bibitem{Breakfast}
Hilde Kuehne, Ali Arslan, and Thomas Serre.
\newblock The language of actions: Recovering the syntax and semantics of
  goal-directed human activities.
\newblock In {\em Proceedings of the IEEE conference on computer vision and
  pattern recognition}, pages 780--787, 2014.

\bibitem{GTEA}
Alireza Fathi, Xiaofeng Ren, and James~M Rehg.
\newblock Learning to recognize objects in egocentric activities.
\newblock In {\em CVPR 2011}, pages 3281--3288. IEEE, 2011.

\bibitem{EPIC-KITCHENS}
Dima Damen, Hazel Doughty, Giovanni~Maria Farinella, Antonino Furnari,
  Evangelos Kazakos, Jian Ma, Davide Moltisanti, Jonathan Munro, Toby Perrett,
  Will Price, et~al.
\newblock Epic-kitchens-100.
\newblock {\em International Journal of Computer Vision}, 130:33--55, 2022.

\bibitem{MviT}
Haoqi Fan, Bo~Xiong, Karttikeya Mangalam, Yanghao Li, Zhicheng Yan, Jitendra
  Malik, and Christoph Feichtenhofer.
\newblock Multiscale vision transformers.
\newblock In {\em Proceedings of the IEEE/CVF International Conference on
  Computer Vision}, pages 6824--6835, 2021.

\bibitem{Omnivore}
Rohit Girdhar, Mannat Singh, Nikhila Ravi, Laurens van~der Maaten, Armand
  Joulin, and Ishan Misra.
\newblock Omnivore: A single model for many visual modalities.
\newblock In {\em Proceedings of the IEEE/CVF Conference on Computer Vision and
  Pattern Recognition}, pages 16102--16112, 2022.

\bibitem{Tokenlearner}
Michael~S Ryoo, AJ~Piergiovanni, Anurag Arnab, Mostafa Dehghani, and Anelia
  Angelova.
\newblock Tokenlearner: What can 8 learned tokens do for images and videos?
\newblock {\em arXiv preprint arXiv:2106.11297}, 2021.

\bibitem{ResT}
Chung-Ching Lin, Kevin Lin, Lijuan Wang, Zicheng Liu, and Linjie Li.
\newblock Cross-modal representation learning for zero-shot action recognition.
\newblock In {\em Proceedings of the IEEE/CVF Conference on Computer Vision and   Pattern Recognition}, pages 19978--19988, 2022.
  

\end{thebibliography}

\end{document}